\documentclass[journal]{IEEEtran}
\usepackage{amssymb}
\usepackage{cite}
\usepackage{amsmath}
\ifCLASSINFOpdf
\usepackage[pdftex]{graphicx}
\usepackage{tabularx}
\usepackage{bm}
\usepackage{url}
\usepackage{xcolor}
\usepackage{soul}

\usepackage{array}
\usepackage{booktabs}
\usepackage{longtable}
\usepackage{pdflscape}

\else
\fi
\usepackage[caption=false,font=normalsize,labelfont=sf,textfont=sf]{subfig}
\begin{document}
\title{MorphoGP: A Nonparametric Framework for Predicting Equilibrium Beach Profiles Under Tidal Influence}
\author{Xi Wu$^{*}$, Yanqing Wei$^{*}$, Hang Yin, Pengze Li, Hongshuai Qi$^{\dagger}$, and Xi Chen$^{\dagger}$,~\IEEEmembership{Member,~IEEE}
\thanks{This work was supported in part by the National Science and Technology Basic Resource Survey Special Project of China under Grant 2022FY202400 and in part by the Ministry of Natural Resources--Provincial Government Joint Science and Technology Project under Grant 2024ZRBSHZ109.}
\thanks{$^{*}$Xi Wu and Yanqing Wei contributed equally to this work.}
\thanks{$^{\dagger}$Hongshuai Qi and Xi Chen are the corresponding authors.}
\thanks{Xi Wu, Pengze Li, and Xi Chen are with the Artificial Intelligence Innovation and Incubation Institute, Fudan University, Shanghai 201203, China, and also with the Shanghai Academy of AI for Science (SAIS), Shanghai 200032, China (e-mail: wux24@m.fudan.edu.cn; lipz25@m.fudan.edu.cn; x\_chen@fudan.edu.cn).}
\thanks{Yanqing Wei is with the State Key Laboratory of Estuarine and Coastal Research, East China Normal University, Shanghai 200062, China (e-mail: 52293904041@stu.ecnu.edu.cn).}
\thanks{Hang Yin is with the College of Ocean and Earth Sciences, Xiamen University, Xiamen 361104, China, and also with the Third Institute of Oceanography, Ministry of Natural Resources, Xiamen 361005, China (e-mail: yinhang@stu.xmu.edu.cn).}
\thanks{Hongshuai Qi is with the Laboratory of Ocean and Coast Geology, Third Institute of Oceanography, Ministry of Natural Resources, Xiamen 361005, China (e-mail: qihongshuai@tio.org.cn).}}
\markboth{IEEE TRANSACTIONS ON GEOSCIENCE AND REMOTE SENSING}%
{Wu \MakeLowercase{\textit{et al.}}: MorphoGP}

\maketitle

\begin{abstract}
The prediction of equilibrium beach profiles under tidal influence is of fundamental importance for sustainable coastal development, informing shoreline protection strategies and managing coastal ecosystems under changing environmental conditions. However, it remains challenging due to the highly nonlinear interactions among wave, tide, and sedimentary processes. Traditional empirical and numerical models often exhibit limited adaptability across diverse coastal environments, with especially pronounced limitations in beach systems where tidal processes are important . To improve data-driven prediction under these conditions, this study proposes MorphoGP, a unified category-specific Gaussian process framework for predicting equilibrium beach profiles (EBPs) under tidal influence. The framework first introduces a ContourCluster model based on contrastive learning to classify tide-influenced beach morphologies automatically. Within each morphological category, a specialized Gaussian process expert learns statistical associations between environmental descriptors including waves, tides, and sediments and the beach profile's shape. A Gating Net then integrates the outputs of all experts through a probabilistic weighting mechanism to produce the final prediction. It should be noted that MorphoGP is a data-driven predictive framework and does not explicitly resolve the full wave--tide--sediment transport dynamics. Instead, it incorporates physically relevant descriptors to support prediction and model interpretation. Evaluated on data from over 180 beach profiles from tide-influenced coasts along the Chinese coast, MorphoGP achieves improved predictive performance compared with conventional and deep learning models, reducing the test RMSE by about 59.3\% compared with the best baseline and achieving a final RMSE of 0.297 m. Additionally, feature relevance analysis suggests that tidal parameters are strongly associated with equilibrium morphology, alongside wave and sediment characteristics. The proposed framework provides a physically informed, data-driven tool for equilibrium beach-profile prediction under tidal influence and coastal management, while stronger process-level physical coupling remains an important direction for future development. The source code is available at \url{https://github.com/Ch1hyaAnon/MorphoGP.git}.
\end{abstract}


\IEEEpeerreviewmaketitle

\section{Introduction}
\IEEEPARstart{E}{quilibrium} beach profiles~\cite{dean1991} provide an idealized but useful framework for representing the long-term adjustment of beach morphology to  hydrodynamic forcing and sediment transport processes~\cite{johnson1919,cornaglia1889}. Their development is influenced by local factors such as wave climate, sediment supply, and geological-geomorphological setting. Because these factors vary in time and space, natural beaches are generally in continuous adjustment rather than fixed equilibrium. Therefore, equilibrium profiles should be regarded as theoretical constructs rather than universally applicable representations of beach morphology. Nevertheless, they are widely used to assess shoreline stability, evaluate beach nourishment performance, and examine coastal resilience under changing environmental conditions. Accurate prediction of equilibrium beach profiles is essential for understanding coastal morphodynamics and informing effective coastal management. While substantial progress has been made in modeling equilibrium profiles for wave-dominated beaches, the applicability of existing approaches to tide-influenced environments remains limited.

In particular, macrotidal beaches, as representative tide-influenced systems, characterized by large tidal ranges and strong tidal currents~\cite{short1991macro}, exhibit complex and highly variable morphological responses~\cite{wright1982morphodynamics} that differ fundamentally from those observed on microtidal coasts. These environments are widespread, occurring along extensive coastlines in northwestern Europe, East Asia, and parts of Australia, and they are socioeconomically important because of their roles in coastal protection, ecosystem services, and human activities. Despite this importance, tidal effects have received comparatively less attention in equilibrium-profile research, leaving a critical gap between scientific understanding and practical modeling needs.

Traditional equilibrium beach profile (EBP) models are largely rooted in parametric or semi-empirical formulations that emphasize wave-driven processes. Classic theories, such as the Dean profile and its extensions, relate profile shape to wave energy dissipation and sediment characteristics, under the implicit assumption that wave forcing dominates morphological adjustment~\cite{dean1991,bruun1954,dean1977,bodge1992,lee1994,larson1999,stokes2015observation}. These approaches have demonstrated considerable success for wave-dominated beaches, where tidal effects are relatively weak or can be treated as secondary influences.

However, the assumptions underlying conventional models are often violated when tidal processes become important~\cite{wei2025analysis,anthony2004morphodynamics}. For example, on macrotidal beaches, tidal currents and wave--tide interactions can play leading roles in sediment transport and profile evolution, giving rise to multiple equilibrium-like states under similar wave conditions. The resulting profiles are often heterogeneous and multimodal, reflecting the coexistence of distinct morphological regimes controlled by tidal phase, sediment redistribution pathways, and hydrodynamic coupling. Consequently, single-regime parametric models struggle to represent the diversity and uncertainty inherent in beach profiles affected by tidal processes.

In response to the limitations of traditional formulations, machine learning (ML) and deep learning (DL) techniques have increasingly been applied to coastal morphodynamic modeling. Data-driven approaches, including neural networks, random forests, and convolutional architectures, have shown promising performance in predicting shoreline change, beach states, and profile evolution by learning complex nonlinear relationships directly from observations~\cite{hashemi2010,de2024,khan2024}. These methods reduce reliance on explicit physical parameterization and can leverage large, heterogeneous datasets.

Recent studies have also made important progress in improving the physical relevance and reliability of ML-based coastal predictions. For example, some studies have incorporated wave-, tide-, and sediment-related descriptors into data-driven models, while others have adopted probabilistic or Gaussian-process-based formulations to provide uncertainty-aware predictions \cite{beuzen2019,adusumilli2024,demelo2024,muroi2025}. Nevertheless, the scope of many existing ML- and DL-based studies is still often limited in ways that hinder their applicability to beach-profile prediction under tidal influence. Rather than being trained and validated across diverse coastal settings, a substantial portion of prior work focuses on time-series prediction using a small number of beaches and a few repeatedly surveyed profiles, which provides limited coverage of morphological variability and hydrodynamic forcing conditions. More importantly, existing uncertainty-aware or physically informed ML models have not fully addressed the regime heterogeneity of equilibrium beach profiles under tidal influence, where different morphological categories may respond differently to wave--tide--sediment forcing. When such models are transferred to tide-influenced coasts, they often face challenges in simultaneously representing morphological regimes, quantifying predictive uncertainty, and maintaining interpretability across heterogeneous sites. In particular, modeling all beach profiles affected by tidal forcing as samples from a single statistical distribution can obscure the presence of multiple morphological regimes driven by distinct wave--tide interactions, yielding overly smoothed profile estimates and reducing generalization across sites and environmental conditions. Although probabilistic ML methods can provide uncertainty estimates, a single global model may still produce poorly resolved uncertainty when the underlying beach profiles are drawn from multiple morphological regimes. Therefore, a category-specific probabilistic framework is needed to combine morphological regime identification, uncertainty-aware prediction, and feature relevance analysis for equilibrium beach-profile prediction under tidal influence.

These limitations highlight the need for a modeling framework that explicitly accounts for the heterogeneous, multi-regime nature of equilibrium beach profiles under tidal influence while retaining predictive flexibility and uncertainty awareness. In this study, the term ``multiple equilibrium states'' refers to the coexistence of distinct, observed equilibrium-profile types under different combinations of wave, tide, and sediment conditions, rather than to a single universal profile form. In representative macrotidal settings, beaches may exhibit different quasi-stable profile geometries, such as relatively steep and narrow profiles, wide low-gradient intertidal profiles, or multi-slope profiles with pronounced cross-shore variability, depending on tidal range, wave exposure, sediment properties, and antecedent morphodynamic conditions. Rather than relying on a single global model or restrictive parametric assumptions, such a framework should be capable of identifying observed morphological regimes in the data, learning localized relationships within distinct regimes, and integrating these components into a coherent probabilistic prediction.

To this end, we propose a nonparametric framework for equilibrium beach-profile prediction under tidal influence. By combining morphology-driven clustering, category-specific Gaussian process experts, and probabilistic model averaging, the proposed approach directly addresses the challenges posed by tidal influence, regime diversity, and predictive uncertainty. Compared with conventional Gaussian process or ensemble models that typically learn a global input--output mapping, MorphoGP improves physical interpretability by conditioning the prediction on observed profile types. Each Gaussian process expert is associated with a morphology-derived regime, and the probabilistic gating weights indicate the relative contribution of different regimes to the final prediction. This design allows the relationships between environmental forcing features and equilibrium-profile geometry to be interpreted within specific morphological contexts, instead of being averaged over all beach types.

The main contributions of this work are threefold:
\begin{enumerate}
  \item We present the first large-scale, data-driven investigation for tide-influenced beaches along the Chinese coast, covering over 180 beaches, including macrotidal, mesotidal, and microtidal sites, and compile a unified dataset that supports systematic model development and evaluation. The dataset enables the observed diversity of tide-influenced equilibrium-profile types to be analyzed in a consistent cross-site framework.

  \item To account for heterogeneous tide-influenced beach morphodynamics, we develop a category-specific nonparametric modeling framework based on Gaussian process experts, enabling flexible and localized equilibrium-profile prediction across distinct morphological regimes. Unlike a single global Gaussian process or a generic ensemble predictor, the proposed framework first identifies morphology-derived profile categories and then learns regime-conditioned forcing--profile relationships, thereby linking predictive components to physically interpretable beach-profile types.

  \item We analyze the correlations between environmental forcing features and equilibrium beach-profile predictions. This analysis highlights the pronounced influence of  tide-related factors in equilibrium beach-profile prediction, underscoring the limitations of wave-centric equilibrium-profile models. By conducting feature relevance analysis within a probabilistic, category-specific framework, MorphoGP provides more interpretable evidence on how wave-, tide-, and sediment-related descriptors are associated with different equilibrium-profile regimes.
\end{enumerate}

The remainder of this article is organized as follows. Section~\ref{relatedworks} reviews related work; Section~\ref{method} presents the proposed MorphoGP framework; Section~\ref{experiment} describes the experimental setup and results; and Section~\ref{conclusion} concludes the paper.

\section{Related Works}
\label{relatedworks}
\subsection{Progress in Equilibrium Beach Profile Research}
The EBP not only reflects the long-term effects of hydrodynamics but also contains information on sediment properties and geomorphic erosion and deposition~\cite{anthony2002between,karunarathna2016linkages}. Early empirical models, notably Bruun (1954) and Dean (1991), established power-law formulations to describe cross-shore profile geometry, primarily based on data from microtidal to mesotidal environments where wave action dominates. Subsequent refinements, such as the exponential model proposed by Bodge (1992) and the segmented equilibrium profile by Larson et al. (1999), further advanced the analytical representation of EBP under varying wave conditions. However, these classical frameworks largely overlooked the role of tides, particularly in macrotidal settings where tidal ranges exceed 4 m and exert comparable or greater morphodynamic influence on beaches.

In recognition of this gap, several studies began to incorporate tidal forcing into beach-state classification. Masselink and Short (1993) proposed the $\Omega$--RTR framework, which combines the dimensionless fall velocity ($\Omega$) with the relative tidal range (RTR) to extend the reflective--dissipative continuum to tidal environments. These morphodynamic studies provide an instructive qualitative taxonomy of beach-profile morphologies; however, they do not offer a quantitative expression for the EBP. More importantly, the predictive skill of the $\Omega$--RTR approach tends to deteriorate in macrotidal settings, where strong tidal currents and wave--tide coupling promote tide-modulated, spatially heterogeneous morphodynamics that depart from the wave-dominated assumptions embedded in the model. Such limitations have been documented along the Shandong Peninsula and the western Guangdong coasts in China~\cite{zhang2023classification,ding2020beach}. Consistently, megatidal beaches in northern France (e.g., Merlimont) exhibit persistent intertidal bar--trough systems and pronounced tidal modulation of sediment transport~\cite{anthony2004morphodynamics}, conditions that further complicate the mapping between $\Omega$--RTR parameters and observed beach states. Recent field observations along the western coast of the Taiwan Strait likewise demonstrate that existing classification schemes struggle to capture terrace-type and dissipative-type profiles shaped by macrotidal forcing, underscoring the need for more flexible, data-driven approaches for quantitative EBP characterization in tide-influenced environments \cite{wei2025analysis}.

\subsection{Process-Based Numerical Beach Morphology Models}
Process-based numerical models have been extensively developed to simulate coastal morphodynamics by explicitly resolving the physical processes governing wave--current interactions, sediment transport, and energy dissipation.
DELFT3D \cite{lesser2004development} implemented and validated sediment transport formulations that allow for coupled simulation of hydrodynamics and sediment transport, making it one of the most widely used coastal process models.
CSTM-ROMS \cite{warner2008development} extended the Regional Ocean Modeling System (ROMS) \cite{haidvogel2008ocean} by implementing three-dimensional sediment transport modeling that accounts for multiple noncohesive sediment classes and dynamic bed morphology.
XBeach \cite{roelvink2010xbeach} was developed to simulate nearshore hydrodynamics and morphological changes, providing insight into wave run-up, overwash, and dune erosion processes under storm conditions.
COAST2D \cite{du2010modelling} was designed to model cross-shore and longshore sediment transport, enabling the prediction of shoreline evolution driven by combined wave and current. These process-based models have been widely applied to simulate coastal morphodynamics and beach profile evolution.

However, these process-based models are constrained by substantial computational demands, extensive parameterization needs, and reliance on detailed initial and boundary conditions \cite{kroon2025parameter}. More critically, their ability to accurately predict equilibrium profiles diminishes in environments with strong spatial heterogeneity and tidal action, especially along macrotidal coasts. The complexity of representing macrotidal ranges, sediment redistribution under combined wave-current actions, and varying sediment supplies hinders their generalizability across diverse and data-sparse settings. These limitations underscore the need for alternative approaches that can leverage growing observational datasets and offer greater adaptability. Data-driven machine learning methods present a compelling pathway, as they can learn complex morphodynamic relationships directly from data, require fewer explicit physical assumptions, and provide scalable predictions with computational efficiency—addressing key gaps left by traditional numerical models.

\subsection{Machine Learning and Deep Learning for Beach Profile Prediction}
With the advent of deep learning, data-driven methods have emerged as powerful alternatives to traditional modeling approaches for predicting beach profile evolution. Before the widespread adoption of deep learning--based spatiotemporal representation learning, Hsu et al. \cite{hsu1986two} introduced a two-dimensional empirical eigenfunction model—an early machine learning approach—for modeling beach profile changes. Subsequently, Hashemi et al. \cite{hashemi2010} employed an Artificial Neural Network (ANN) to predict seasonal beach profile variations along Tremadoc Bay, using beach geometry, wind data, and local wave conditions as input features. Later, de Melo et al. \cite{de2024} applied the Random Forest (RF) algorithm to simulate the seasonal morphological evolution of several Portuguese beaches, while Khan et al. \cite{khan2024} utilized Convolutional Neural Networks (CNNs) to predict long-term coastal changes driven by sea-level rise and wave conditions. Disdier et al. \cite{disdier2025predicting} developed an ANN model with three hidden layers to predict beach profiles from offshore wave spectra. Although this model achieved high predictive accuracy, its performance was constrained by reliance on synthetic training data generated from FUNWAVE-TVD modeling, which limited its real-world applicability. More recently, Kim et al. \cite{kim2025prediction} proposed spatiotemporal Graph Neural Networks (GNNs) to improve the prediction of beach morphological evolution by capturing nonlinear relationships and spatiotemporal dependencies among multiple environmental variables, incorporating satellite optical imagery as an additional remote-sensing input.

Despite these advances, most existing studies have focused on time-series predictions for beaches without adequately considering tidal influence on beach processes, and several rely on synthetically generated datasets. These limitations are particularly evident in macrotidal settings, where tides can play an important role in shaping beach morphology. Furthermore, many machine learning models suffer from the “black-box” problem, meaning that they provide accurate predictions but lack interpretability and transparency regarding the relationships between environmental variables and beach profile characteristics.

In contrast, our approach seeks to bridge this gap by utilizing Gaussian Process (GP) \cite{williams2006gaussian}, which provides a more interpretable and probabilistic solution to beach profile prediction. GPs can naturally account for uncertainty and provide insight into the importance of different environmental drivers, making them well suited for modeling the complex and highly nonlinear relationships in equilibrium beach-profile prediction under tidal influence.

\subsection{Unsupervised Morphological Classification and Regime-Specific Ensemble Modelling}

Morphodynamic regime classification has long provided an important basis for interpreting beach-profile variability. A widely used framework for tide-influenced beaches is the Masselink--Short $\Omega$--RTR model, which classifies beach states using the dimensionless fall velocity $\Omega$ and the relative tide range RTR \cite{masselink1993effect}. In this framework, $\Omega$ reflects the relative effects of wave forcing and sediment settling, whereas RTR represents the relative importance of tidal range compared with breaking wave height. By combining these two physically meaningful parameters, the Masselink--Short framework links beach morphology to wave, tide, and sediment conditions and distinguishes morphodynamic states such as reflective, intermediate, dissipative, low-tide terrace, and tide-dominated beach types. It therefore provides an essential physical reference for tide-influenced beach classification.

In addition to parameter-based morphodynamic classifications, shape-based learning methods have provided another important route for profile and contour classification. Shapelet-based methods are particularly relevant because they identify discriminative local subsequences or sub-shapes that are representative of different classes \cite{ye2009shapelets}. By transforming each sequence into distances to a set of informative shapelets, the shapelet transform enables local, phase-independent shape comparison and can improve the interpretability of time-series or profile classification \cite{lines2012shapelet,hills2014classification}. Later studies further extended this idea by learning shapelets directly from data through optimization-based formulations \cite{grabocka2014learning}. For coastal profile analysis, such methods are conceptually attractive because local geometric patterns, such as beachface steepening, terrace-like flattening, bar--trough structures, or changes in profile curvature, may serve as diagnostic indicators of different morphodynamic states.

Nevertheless, most shapelet-based methods are designed primarily for supervised classification or discriminative feature extraction, whereas the present study requires unsupervised discovery of morphology-derived regimes and their integration with downstream probabilistic EBP prediction. ContourCluster is therefore developed not simply to classify profiles, but to learn morphology-aware categories that can define regime-specific Gaussian process experts and support uncertainty-aware prediction across heterogeneous tide-influenced beaches.

However, the $\Omega$--RTR framework was originally designed as a conceptual morphodynamic classification scheme rather than as a morphology-driven representation for profile prediction. Its classes are primarily determined from a small set of hydrodynamic and sediment-related parameters and may not fully capture the geometric diversity of complete cross-shore equilibrium profiles, especially when beaches with similar $\Omega$ and RTR values exhibit different profile curvature, intertidal width, bar/terrace expression, or multi-slope structure. For equilibrium beach-profile prediction, this limitation is important because the subsequent regression model requires profile categories that are not only physically interpretable but also geometrically coherent in the observed profile space.

\begin{figure*}[!ht]
    \centering
    \includegraphics[width=\textwidth]{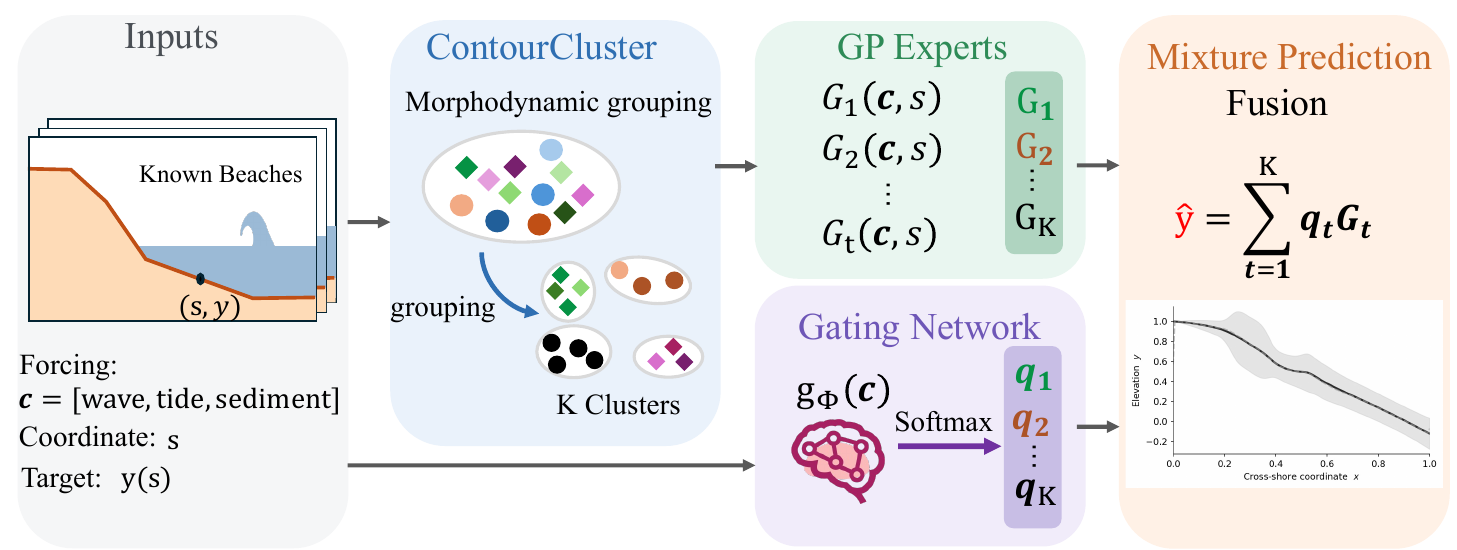}
   \caption{\textbf{Overview of the proposed MorphoGP framework.}
Given environmental forcing descriptors \(\mathbf{c}=[\text{wave},\text{tide},\text{sediment}]\) and the cross-shore coordinate \(s\), MorphoGP predicts the equilibrium beach profile \(y(s)\).
ContourCluster first groups observed profiles into \(K\) morphodynamic clusters.
A set of cluster-specific Gaussian-process experts \(\{G_t(\mathbf{c},s)\}_{t=1}^{K}\) models profile variability within each cluster, while a gating network \(g_{\Phi}(\mathbf{c})\) outputs mixture weights \(\mathbf{q}=\mathrm{softmax}(g_{\Phi}(\mathbf{c}))\).
The final prediction is obtained by fusing the expert outputs via \(\hat{y}(s)=\sum_{t=1}^{K} q_t\,G_t(\mathbf{c},s)\), yielding an uncertainty-aware profile estimate (inset).}
\label{fig_overview}
\end{figure*}

A growing body of work has explored unsupervised learning techniques to classify beach profiles based on their intrinsic morphological features, thereby making the strong heterogeneity between beach states explicit. Classical approaches include $k$-means clustering and principal component analysis (PCA), which group profiles according to low-dimensional shape descriptors. For example, Riazi~\cite{riazi2021subaerial} applied unsupervised clustering techniques to 916 subaerial beach profiles collected from six beaches along the Maine coast between 2005 and 2018 and successfully grouped them into multiple categories based on their shape characteristics. This study demonstrated the potential of data-driven methods to capture the inherent variability of beach morphologies beyond traditional parameter-based classifications.

Once distinct morphological regimes have been identified, it is natural to model each regime with a specialized predictor rather than enforcing a single global relationship for all profiles. In statistics and machine learning, this idea appears in various forms such as piecewise regression \cite{toms2003piecewise}, latent-class or regime-switching models \cite{hamilton1989new}, and ensemble methods \cite{dietterich2000ensemble} in which several sub-models are trained on different parts of the input space and their predictions are combined through data-dependent weights. Such heterogeneity-aware frameworks have been shown to improve both predictive performance and interpretability in complex systems with multiple operating states.

For coastal morphodynamics, this perspective is particularly appealing: different beach states, including reflective, dissipative, low-tide terrace, and tide-dominated morphologies \cite{masselink1993effect}, are associated with distinct combinations of wave, tide, and sediment conditions and may obey different effective morphodynamic relationships. Nevertheless, a key unresolved question is whether a conceptual parameter-based classification such as the Masselink--Short $\Omega$--RTR scheme provides sufficiently coherent categories for data-driven equilibrium-profile prediction, or whether profile-shape-based clustering can better represent the observed geometric structure of macrotidal EBPs. However, the combination of unsupervised morphology-based classification with category-specific probabilistic regression models has not yet been systematically explored for equilibrium beach-profile modeling, especially under macrotidal conditions. In this work, we address this gap by integrating a contour-based unsupervised classifier with a set of Gaussian Process models and a probabilistic weighting mechanism that selects and combines them according to the inferred morphodynamic category, enabling morphology-aware, uncertainty-quantified EBP prediction across heterogeneous tide-influenced beaches. To directly assess the novelty and utility of the clustering module, we further compare ContourCluster with the standard Masselink--Short $\Omega$--RTR classification in terms of both morphological consistency and downstream prediction performance.

\section{Methodology}
\label{method}

\subsection{Overview}

\noindent\textbf{Problem setup and notation.}
Let $\mathcal{D}=\{(\mathbf{c}_i,\mathbf{y}_i)\}_{i=1}^{N}$ denote a dataset of $N$ surveyed equilibrium beach profiles.
For profile $i$, $\mathbf{c}_i\in\mathbb{R}^{d}$ collects the environmental drivers (wave, tide, and sediment descriptors), and
$\mathbf{y}_i\in\mathbb{R}^{M}$ is the corresponding profile sampled on a fixed cross-shore grid $\{s_m\}_{m=1}^{M}$, i.e.,
$\mathbf{y}_i=[y_i(s_1),\ldots,y_i(s_M)]^\top$, where $y_i(s_m)$ denotes bed elevation (or depth) at coordinate $s_m$ under a consistent vertical datum.
For notational simplicity, we drop sample indices and write $\mathbf{y}$ and $\mathbf{c}$ for generic profile and forcing vectors.
Given a new forcing vector $\mathbf{c}_*$, our goal is to predict the conditional distribution
\begin{equation}
p(\mathbf{y}\mid \mathbf{c}),
\end{equation}
and to provide both a point estimate (e.g., posterior mean profile) and a coherent predictive uncertainty.
In implementation, MorphoGP predicts each ordinate $y(s_m)$ using the augmented input $\mathbf{z}=[\mathbf{c},\,s_m]$ (thus $D=d+1$)
and assembles the full profile by evaluating the model for $m=1,\ldots,M$.

\vspace{0.25em}
This work introduces MorphoGP, a framework for predicting equilibrium beach profiles under tidal influence from key environmental drivers,
including wave, tide, and sediment characteristics. As illustrated in Fig.~\ref{fig_overview}, the framework consists of three components:
(1) a ContourCluster module that assigns each profile to one of $K$ morphology categories based on contour-derived shape features;
(2) $K$ category-specific Gaussian process (GP) experts trained to model $p\!\left(y(s)\mid \mathbf{c},s,\,\kappa=k\right)$ for each category, where $\kappa\in\{1,\ldots,K\}$ denotes the morphology category;
and (3) a gating network that maps $\mathbf{c}$ to input-dependent mixture weights and combines the expert predictive distributions into
$p(\mathbf{y}\mid \mathbf{c})$ with uncertainty quantification. When encountering a new beach characterized by $\mathbf{c}_*$,
MorphoGP evaluates the gating network to obtain expert weights and produces the final profile prediction by aggregating expert outputs across all $s_m$.

The following sections detail these components.

\subsection{ContourCluster Model}
\label{Contour}
To establish objective, data-driven domains for the expert models, we categorize beach profiles based on morphological features derived from their cross-shore elevation sequences $\mathbf{y}_i$.
The ContourCluster module comprises two stages: (i) a beach-shapelet extractor that embeds profile subsequences into a discriminative representation space via contrastive learning and (ii) $k$-means clustering to obtain $K$ morphology categories.

\begin{figure}[!h]
\centering
\includegraphics[width=3.2in]{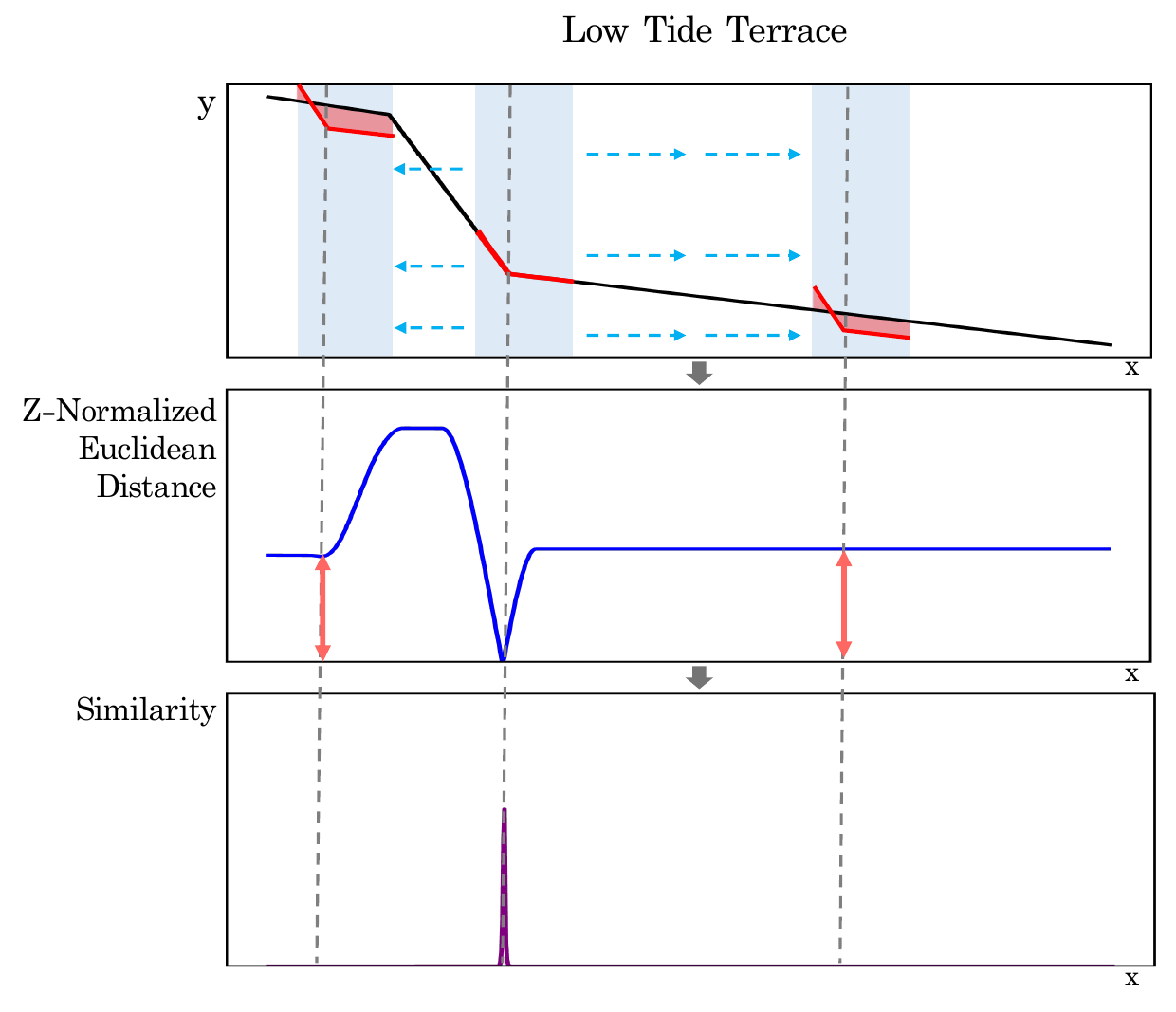}
\caption{Illustration of shapelet-based matching on a representative \emph{Low Tide Terrace} equilibrium beach profile \cite{masselink1993effect}. \textbf{Top:} a discriminative subsequence (shapelet) \(S\) highlighted on the reference profile \(r(x)\). \textbf{Middle:} z-normalized Euclidean distance profile \(d_u\) obtained by sliding a window \(w_u\) of length \(L\) along \(r\). \textbf{Bottom:} normalized similarity weights \(\chi_u\) computed via the soft-min transform in~\eqref{eq:shapelet_softmin}, where larger \(\chi_u\) indicates higher local morphological similarity.}
\label{shapelets}
\end{figure}

This work leverages shapelets~\cite{grabocka2014learning}, which are discriminative subsequences of time series that capture local patterns and serve as interpretable features for classification.
As shown in Fig.~\ref{shapelets}, the highlighted subsequence illustrates a shapelet $S$, and its distance profile is obtained by sliding a window $w_u$ across a reference series $r$.
Before computing the distance, both the shapelet and each window are z-normalized~\cite{rakthanmanon2013addressing} to eliminate the influence of amplitude differences:
\begin{equation}
    S' = \frac{S - \operatorname{mean}(S)}{\operatorname{std}(S)}, \quad
    w_u' = \frac{w_u - \operatorname{mean}(w_u)}{\operatorname{std}(w_u)} .
\end{equation}
Let $S'\in\mathbb{R}^{L}$ and $w_u'\in\mathbb{R}^{L}$ denote the z-normalized subsequences of length $L$.
The normalized Euclidean distance is computed as
\begin{equation}
\label{eq:shapelet_euc}
    d_u = \sqrt{\sum_{j=1}^{L} (S'_j - w'_{u,j})^2} .
\end{equation}
To emphasize regions of highest morphological similarity, the distances $\{d_u\}$ are transformed into normalized similarity weights $\{\chi_u\}$ using a soft-min function:
\begin{equation}
\label{eq:shapelet_softmin}
    \chi_u = \frac{\exp(-\gamma d_u)}{\sum_{u'=1}^{N_w} \exp(-\gamma d_{u'})},
\end{equation}
where $\gamma>0$ is a temperature parameter and $N_w$ is the number of sliding windows.
Finally, $k$-means clustering is applied to group profiles into different categories according to their similarity weights $\{\chi_u\}$ and slope characteristics.

\begin{figure}[htbp]
\centering \includegraphics[width=3.0in]{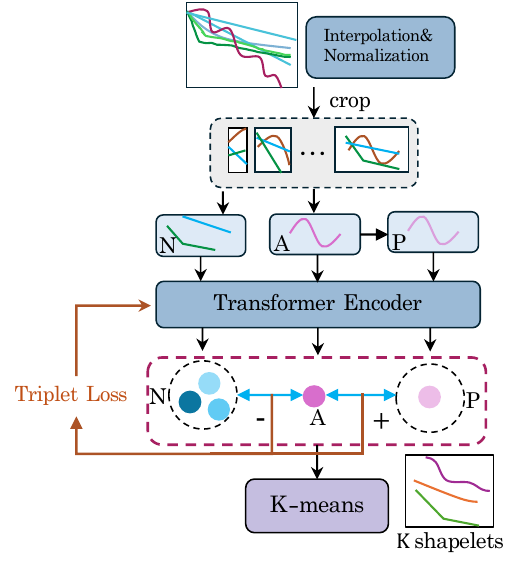}
\caption{Overall workflow for obtaining shapelets. The workflow integrates candidate extraction, a transformer-based embedding layer, and a selection strategy.} \label{shapenet}
\end{figure}

\textbf{Beach Shapelets Extractor.}
We employ contrastive learning to extract discriminative shapelets---i.e., representative and diagnostic segments of the beach profile or outline---from different types of beaches.
The workflow is detailed below and illustrated in Fig.~\ref{shapenet}.

\paragraph{Shapelets Candidate Extraction}
Since the raw beach profile data consist of discrete samples $\{(s_m, y_m)\}_{m=1}^{M}$, we perform linear interpolation to obtain an elevation sequence $u=\{u_n\}_{n=1}^{N_s}$ of fixed length $N_s$.
To ensure comparability across samples, the sequence is rescaled by min--max normalization:
\begin{equation}
    \tilde{u}_n = \frac{u_n - \min(u)}{\max(u) - \min(u)}, \quad n=1,2,\dots,N_s.
\end{equation}
From each normalized profile, subsequences of different lengths $L \in \mathcal{L}=\{L_1, L_2, \dots, L_Q\}$ are cropped as shapelet candidates:
\begin{equation}
    a_{u,L} = \tilde{u}_{u:u+L-1}, \quad u=1,2,\dots,N_s-L+1.
\end{equation}

\begin{figure}[htbp]
\centering
\includegraphics[width=2.5in]{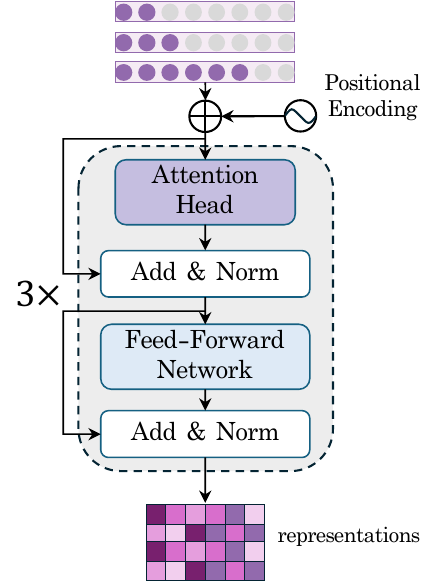} \caption{\textbf{Structure of the Transformer encoder.} The encoder comprises three stacked Transformer layers, each equipped with three multi-head attention modules.}
\label{trans}
\end{figure}

\paragraph{Transformer-based Embedding Layer}
As illustrated in Fig.~\ref{trans}, each shapelet candidate $a_{u,L}$ is mapped into a low-dimensional embedding vector through a Transformer-based encoder~\cite{vaswani2017attention}.
Before entering the encoder, positional encoding is added to retain sequential order. The embedding process is denoted as:
\begin{equation}
    v_{u,L} = f_{\theta}(a_{u,L}),
\end{equation}
where $f_{\theta}(\cdot)$ is the Transformer encoder parametrized by $\theta$, and $v_{u,L} \in \mathbb{R}^{d_e}$ is the fixed-size embedding vector of dimension $d_e$.

\paragraph{Contrastive Training with Triplet Loss}

To construct a discriminative embedding space, we employ contrastive learning with triplet loss~\cite{hermans2017defense}. 

Triplet construction is guided by local geometric descriptors computed from each shapelet candidate. Positive and negative samples are defined based on similarity in this descriptor space, rather than purely random sampling or augmentation alone.

For each candidate shapelet \(a_{u,L}\), we compute a local morphology descriptor
\begin{equation}
    \mathbf{g}_{u,L} =
    \left[
    \bar{x}_{u,L},\,
    \bar{z}_{u,L},\,
    \mu_{\nabla y}^{u,L},\,
    \sigma_{\nabla y}^{u,L},\,
    \mu_{\kappa}^{u,L},\,
    \sigma_{\kappa}^{u,L}
    \right],
\end{equation}
where the elements describe the normalized cross-shore location, relative elevation, and statistical properties of local slope and curvature within the subsequence. This descriptor provides a compact representation of local geometric structure.

Positive samples are defined as (i) augmentation-based variants generated by weak perturbations that preserve local structure, and (ii) descriptor-nearest neighbors satisfying
\begin{equation}
    D_g(a_{u,L},a_{p,L_p}) < \tau_p,
\end{equation}
where \(D_g(\cdot,\cdot)\) measures Euclidean distance in the standardized descriptor space and \(\tau_p\) is a threshold.

Negative samples are selected from candidates with sufficiently different descriptor values, defined as
\begin{equation}
    D_g(a_{u,L},a_{n,L_n}) > \tau_n,
\end{equation}
where \(\tau_n > \tau_p\). Semi-hard negatives are also included when their embedding similarity is high but descriptor distance exceeds \(\tau_n\), improving robustness of the embedding space.

Let \(v_a = f_{\theta}(a_{u,L})\), \(v_p = f_{\theta}(a_p)\), and \(v_n = f_{\theta}(a_n)\).
The triplet loss is defined as:
\begin{equation}
    \mathcal{L}_{\text{triplet}} =
    \max \Big( 0,\,
    \lVert v_a - v_p \rVert_2^2
    - \lVert v_a - v_n \rVert_2^2 + m \Big),
\end{equation}
where \(m>0\) is a margin hyperparameter.

The overall training objective is computed by averaging over all valid triplets in a mini-batch:
\begin{equation}
    \mathcal{L} = \frac{1}{|\mathcal{T}|}
    \sum_{(a,p,n)\in \mathcal{T}}
    \mathcal{L}_{\text{triplet}}(a,p,n),
\end{equation}
where \(\mathcal{T}\) denotes the set of descriptor-guided triplets in the batch. Compared with random sampling, this strategy introduces a structured constraint on triplet formation based on local geometric descriptors.

\paragraph{Cluster and Choose}
After obtaining the embedding representations $\{v_{u,L}\}$ of all shapelet candidates, we apply $k$-means to group them into $K$ clusters based on morphological similarity in the embedding space.
Formally, the clustering objective is:
\begin{equation}
    \min_{\{C_k\}_{k=1}^{K}} \sum_{k=1}^{K} \sum_{v_{u,L} \in C_k} \lVert v_{u,L} - \varepsilon_k \rVert_2^2,
\end{equation}
where $C_k$ denotes the $k$-th cluster and $\varepsilon_k$ is its centroid.
From each cluster, we select the shapelet that is closest to the centroid as the representative shapelet:
\begin{equation}
    s_k^\ast = \arg \min_{a_{u,L} \in C_k} \lVert f_{\theta}(a_{u,L}) - \varepsilon_k \rVert_2, \quad k=1,2,\dots,K.
\end{equation}
The final set of discriminative shapelets is
\begin{equation}
    \mathcal{S}^\ast = \{ s_1^\ast, s_2^\ast, \dots, s_K^\ast \}.
\end{equation}

\paragraph{Post-hoc Projection and Geomorphic Interpretation of Learned Shapelets}

To further examine the learned shapelets, we perform a post-hoc projection analysis after the shapelet selection step. This analysis is not used in model training and is intended to visualize where each shapelet best matches observed cross-shore profiles. For each representative shapelet \(s_k^\ast\), we slide it along each normalized profile and compute the similarity weight \(\chi_{i,k,u}\) between \(s_k^\ast\) and the profile window starting at cross-shore position \(u\) on the \(i\)-th profile. The best-matching location is identified as
\begin{equation}
    u_{i,k}^{\ast} = \arg\max_{u} \chi_{i,k,u},
\end{equation}
where \(u_{i,k}^{\ast}\) denotes the location with maximum similarity between the shapelet and the observed profile segment. The matched segment is then mapped back to the original elevation--distance coordinate system for visualization.

We then summarize the local geometric properties of the matched segments, including relative cross-shore position, elevation variation, slope, and curvature. These descriptors are used to characterize the geometric patterns captured by each shapelet. For example, some matched segments correspond to low-gradient regions, while others exhibit sharper slope transitions or convex--concave structures along the profile.

These descriptions should be interpreted as geometric pattern characterizations rather than explicit geomorphological or process-based interpretations. The goal of this analysis is to provide an intuitive understanding of the local profile structures encoded by the shapelets, rather than to assign definitive physical labels or infer underlying sediment transport mechanisms.

\subsection{Category-Specific Experts}
The Category-Specific Experts consist of $K$ independent regression models.
Expert $t\in\{1,\ldots,K\}$ is responsible for predicting profile elevations for the $t$-th morphology category inferred in Section~\ref{Contour}.
We use Gaussian processes (GPs)~\cite{williams2006gaussian} with automatic relevance determination (ARD)~\cite{mackay1992bayesian}
to model the mapping from environmental drivers to profile elevation while quantifying predictive uncertainty.

For a given cross-shore coordinate $s$ and forcing vector $\mathbf{c}\in\mathbb{R}^{d}$, we define the augmented input
\begin{equation}
\mathbf{z}=[\mathbf{c},\, s]\in\mathbb{R}^{D}, \qquad D=d+1.
\end{equation}
The $t$-th expert models the elevation as
\begin{equation}
y = g_t(\mathbf{z}) + \epsilon, \qquad \epsilon\sim\mathcal{N}(0,\sigma^2),
\end{equation}
with GP prior
\begin{equation}
g_t(\mathbf{z}) \sim \mathcal{GP}\!\big(m(\mathbf{z}),k_t(\mathbf{z},\mathbf{z}')\big),
\end{equation}
where $m(\mathbf{z})$ is set to zero and $k_t(\cdot,\cdot)$ is the covariance function.
We adopt an ARD squared-exponential kernel:
\begin{equation}
k_{\text{ARD}}(\mathbf{z}, \mathbf{z}')
= \sigma_f^2 \exp\!\left(-\tfrac{1}{2}(\mathbf{z}-\mathbf{z}')^\top \Lambda^{-1}(\mathbf{z}-\mathbf{z}')\right),
\end{equation}
where $\Lambda=\mathrm{diag}(\ell_1^2,\ldots,\ell_D^2)$ and $\ell_d$ is the length scale for the $d$-th input dimension.
Given training data $\mathcal{D}_t=\{(\mathbf{z}_i,y_{t,i})\}_{i=1}^{n_t}$ from category $t$, the predictive distribution at $\mathbf{z}_*$ is Gaussian,
\begin{equation}
p(y_*\mid \mathbf{z}_*,\mathcal{D}_t,M_t)=\mathcal{N}\!\big(\mu_t(\mathbf{z}_*),\sigma_t^2(\mathbf{z}_*)\big),
\end{equation}

where $\mu_t(\cdot)$ and $\sigma_t^2(\cdot)$ are given by the standard GP posterior.
Hyperparameters $\theta_t=\{\sigma_f^2,\{\ell_d\}_{d=1}^{D},\sigma^2\}$ are learned by maximizing the log marginal likelihood (LML).
Let $\mathbf{Z}_t=[\mathbf{z}_1,\ldots,\mathbf{z}_{n_t}]^\top\in\mathbb{R}^{n_t\times D}$ and
$\mathbf{y}_t=[y_{t,1},\ldots,y_{t,n_t}]^\top\in\mathbb{R}^{n_t}$ denote the training inputs and outputs for expert $t$.
The LML is
\begin{align}
\log p(\mathbf{y}_t \mid \mathbf{Z}_t, \theta_t)
&= -\tfrac{1}{2}\mathbf{y}_t^\top (\mathbf{K}_t + \sigma^2 \mathbf{I})^{-1} \mathbf{y}_t \notag \\
&\quad -\tfrac{1}{2}\log \lvert \mathbf{K}_t + \sigma^2 \mathbf{I} \rvert
-\tfrac{n_t}{2}\log 2\pi ,
\end{align}
where $(\mathbf{K}_t)_{ij}=k_{\text{ARD}}(\mathbf{z}_i,\mathbf{z}_j)$.

\begin{figure*}[!ht]
\centering
\includegraphics[width=\textwidth]{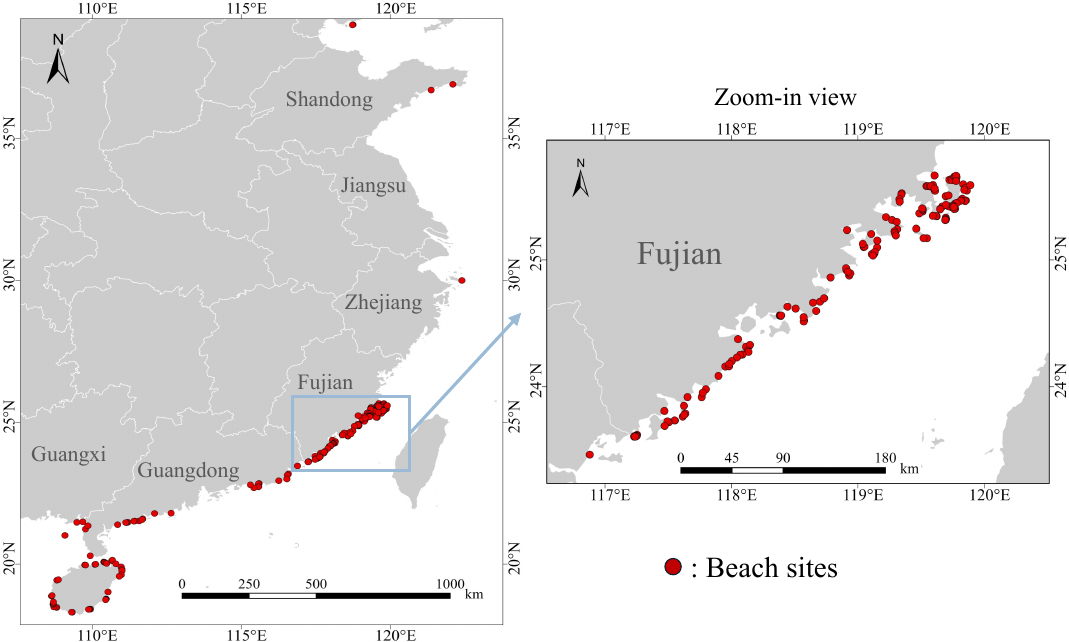}
\caption{Study area and locations of beach-profile surveys along the Chinese coast.} \label{fig:study}
\end{figure*}

\subsection{Gating Net}

To integrate the $K$ category-specific GP experts, we employ a gating network that produces cross-shore-dependent mixture weights.
In the actual implementation, the gating network takes both the environmental forcing vector $\mathbf{c}$ and the cross-shore coordinate $s$ as inputs.
For a given forcing vector $\mathbf{c}$ and cross-shore coordinate $s$, the mixture predictive distribution is
\begin{equation}
p(y\mid s,\mathbf{c})=
\sum_{t=1}^{K}
p(y\mid s,\mathbf{c},M_t)\,q_\phi(t\mid \mathbf{c},s),
\label{eq:moe}
\end{equation}
where each expert $M_t$ provides a Gaussian prediction
$p(y\mid s,\mathbf{c},M_t)=\mathcal{N}\big(y;\mu_t(\mathbf{z}),\sigma_t^2(\mathbf{z})\big)$ with $\mathbf{z}=[\mathbf{c},s]$,
and $q_\phi(t\mid \mathbf{c},s)$ denotes the cross-shore-dependent gating weight.

Motivated by Bayesian model averaging, we parameterize the gating weight by combining an evidence term and a learned contextual term:
\begin{align}
l_t(\mathbf{c},s)&=\beta\,\widetilde{\log p(\mathcal{D}_t \mid M_t)}+g_\phi(\mathbf{c},s)_t,\\
q_\phi(t\mid \mathbf{c},s)&=\frac{\exp(l_t(\mathbf{c},s))}
{\sum_{j=1}^{K}\exp(l_j(\mathbf{c},s))},
\label{eq:bma_softmax}
\end{align}
where $g_\phi(\mathbf{c},s)_t$ is the $t$-th logit output of an MLP with input $[\mathbf{c},s]$,
$\widetilde{\log p(\mathcal{D}_t \mid M_t)}$ denotes a normalized log-evidence term computed from the GP marginal likelihood for expert $t$,
and $\beta$ controls the trade-off between evidence and contextual weighting.
The inclusion of $s$ allows the mixture weights to vary between the upper beach, beachface, intertidal zone, and lower foreshore, thereby reflecting spatially varying morphodynamic controls.

The resulting predictive mean and variance are
\begin{align}
\mathbb{E}[\hat{y}]
&=\sum_{t=1}^{K} q_\phi(t\mid \mathbf{c},s)\,\mu_t(\mathbf{z}),\notag\\
\mathrm{Var}[\hat{y}]
&=\sum_{t=1}^{K} q_\phi(t\mid \mathbf{c},s)
\big(\sigma_t^2(\mathbf{z})+\mu_t^2(\mathbf{z})\big)\notag\\
&\quad-\Big(\sum_{t=1}^{K} q_\phi(t\mid \mathbf{c},s)\,\mu_t(\mathbf{z})\Big)^{2}.
\end{align}

All parameters are learned by maximizing the conditional mixture log-likelihood:
\begin{equation}
\begin{aligned}
\mathcal{L}(\{\theta_t\},\phi)
&=\sum_{i=1}^{N}\sum_{m=1}^{M}
\log\!\Bigg(
\sum_{t=1}^{K} q_\phi(t\mid \mathbf{c}_i,s_m)\,
\mathcal{N}\big(y_i(s_m);\\[-0.2em]
&\hspace{3.2em}\mu_t(\mathbf{z}_{i,m}),\sigma_t^2(\mathbf{z}_{i,m})\big)
\Bigg),
\end{aligned}
\label{eq:bma_ll}
\end{equation}
where $\mathbf{z}_{i,m}=[\mathbf{c}_i,s_m]$, $\{\theta_t\}$ are the GP hyperparameters, and $\phi$ are the gating-network parameters.

This formulation avoids using a single fixed mixture over the entire profile and instead learns a spatially varying expert combination at each cross-shore coordinate. Because the expert predictions are conditioned on the augmented input \(\mathbf{z}=[\mathbf{c},s]\), the cross-shore coordinate is involved in both the GP posterior and the gating weights. As a result, the predicted profile is shaped by two complementary sources of spatial dependence: the covariance-based variation of each GP expert along \(s\), and the gradual change of expert contributions across different cross-shore zones. This design is consistent with the continuous organization of beach-profile morphology, while still allowing local variations associated with berms, terraces, or bar--trough structures.

\section{Experimental Results and Discussion}
\label{experiment}

\subsection{Study area and Data}

\subsubsection{Study Area}
We analyzed 222 cross-shore profiles from 183 sandy beaches along the Chinese coast, distributed along open-coast settings and covering a broad range of hydrodynamic and sedimentary conditions. For each beach, we compiled basic geographic and morphological information to support spatial analysis and to link local morphology with hydrodynamics and sediment characteristics.

\subsubsection{Beach profile and sediment data}
We surveyed beach profiles during the period of lowest tide using a STONEX S9II PRO RTK-GPS. The measured profiles generally extended from the backshore to the low-water level and were referenced to a common elevation datum. Field surveys were conducted between April 2023 and September 2024 using consistent survey procedures across all sites. Although the field surveys included part of the supratidal zone, the present analysis focused specifically on the intertidal portion of the profiles rather than the full active beach profile extending from the frontal dune to the closure depth. The supratidal zone was excluded from the analysis because it is generally weakly affected by normal tidal and wave processes and is typically modified only during extreme events such as storm surges or exceptionally high tides. The subtidal zone was not included because accurate field measurement in submerged environments is strongly constrained by water depth, wave disturbance, and the operational limitations of RTK-GPS surveys. Consequently, the analyzed profiles primarily represent the morphodynamic characteristics of the intertidal beach under tidal influence. In this study, these lowest-tide intertidal profiles are therefore treated as practical approximations of equilibrium-like intertidal beach profiles, rather than as complete long-term averaged equilibrium beach profiles over the entire active cross-shore domain. We acknowledge that true equilibrium beach profiles ideally represent long-term averaged morphology, whereas the profiles used here were measured during individual low-tide surveys. This distinction should be considered when interpreting the predicted equilibrium profiles, as the results represent equilibrium-like intertidal profiles under tidal influence and may not fully capture the complete active beach profile from the frontal dune to the closure depth or longer-term temporal variability.

Surface sediment sampling was carried out synchronously with the profile measurements. For each beach, we selected one or more representative cross-shore profiles based on coastal morphology and sediment distribution, and we preferentially sampled in the mid-tidal zone. At least one surface sample (5--20\,cm depth) was collected per beach, with additional samples where needed to capture alongshore or cross-shore variability. Each sample had a minimum mass of 500\,g.

In the laboratory, we analyzed grain size of the 2023 sediment samples using a POWTEQ SS2000 vibrating sieve shaker. Approximately 50\,g of dried sediment (oven-dried at 60\,$^\circ$C for at least 24\,h) was gently disaggregated when necessary and sieved through a stack of standard sieve meshes (aperture range: 0.063--4.00 mm) at an amplitude of 0.5$\Phi$ until the sediment mass on each sieve reached a stable state. The mass retained on each sieve was weighed and recorded. From the grain-size distributions, we calculated standard statistical parameters such as mean grain size, sorting, skewness, and kurtosis following classical sedimentological procedures \cite{folk1957brazos}.

\begin{table*}[!ht]
\renewcommand{\arraystretch}{1.3}
\caption{Performance comparison of different models on the test set.
Values are reported as mean $\pm$ standard deviation across the five cross-validation folds.
Arrows indicate desirable metric directions (\(\uparrow\): higher is better; \(\downarrow\): lower is better).}
\label{tab:performance}
\centering
\footnotesize
\begin{tabular}{lccc}
\hline
\textbf{Model} & \textbf{\(\bm{R^2}\) (\(\uparrow\))}
 & \textbf{RMSE (\(\downarrow\))} & \textbf{MAE (\(\downarrow\))} \\
\hline
KNN & $0.533 \pm 0.041$ & $0.848 \pm 0.063$ & $0.595 \pm 0.048$ \\
Linear Regression & $0.609 \pm 0.038$ & $0.776 \pm 0.052$ & $0.595 \pm 0.045$ \\
MLP & $0.479 \pm 0.057$ & $0.896 \pm 0.071$ & $0.601 \pm 0.059$ \\
Random Forest & $0.654 \pm 0.035$ & $0.730 \pm 0.049$ & $0.462 \pm 0.037$ \\
SVR & $0.569 \pm 0.044$ & $0.815 \pm 0.058$ & $0.550 \pm 0.046$ \\
Transformer & $0.600 \pm 0.050$ & $0.784 \pm 0.061$ & $0.527 \pm 0.052$ \\
Bruun-type (transformer-based models) & $0.623 \pm 0.039$ & $0.779 \pm 0.054$ & $0.498 \pm 0.041$ \\
Exponential (transformer-based models) & $0.517 \pm 0.046$ & $0.832 \pm 0.067$ & $0.599 \pm 0.055$ \\
\textbf{MorphoGP (Ours)} & $\mathbf{0.942 \pm 0.022}$ & $\mathbf{0.297 \pm 0.041}$ & $\mathbf{0.209 \pm 0.038}$ \\
\hline
\end{tabular}
\end{table*}

\subsubsection{Wave data}

Wave conditions along the study coast were analyzed with TOMAWAC \cite{benoit1996tomawac}. The computational domain was discretized with an unstructured triangular mesh, and bathymetry was derived from the GEBCO global dataset combined with CMAP coastal bathymetric data. The model was applied at 183 nearshore locations, hereinafter referred to as hydrodynamic observation points, which were set in the offshore area adjacent to beaches and can characterize the regional hydrodynamic conditions of the corresponding beaches.

The model performance was systematically validated through multiple approaches, demonstrating reliable accuracy. First, simulations of nine historical typhoon events (e.g., Mangkhut-1822 and Rammasun-1409) were compared with in situ observations from 11 wave stations. The results showed a mean bias in significant wave height of no more than 0.14~m (relative error $<8.8\%$) and a mean bias in wave period of no more than 0.57~s (relative error $<8.6\%$).

These validations confirm the robustness of the TOMAWAC model under both extreme and routine wave conditions. Based on the validated model, time series of significant wave height and period for the year 2023 were obtained at observation points. The maximum significant wave height and period were extracted over 12-hour intervals, and annual mean values of wave parameters were statistically analyzed. The nearshore breaking wave height was calculated (Komar, 1973) and was adopted as the wave descriptor in this study.

\subsubsection{Tidal data}
Tidal conditions at the same 183 points were simulated using the MIKE~21 global tide model, which was validated against long-term tide gauge data from four national marine observation stations (Pingtan, Chongwu, Xiamen, and Dongshan) along the western coast of the Taiwan Strait. The model produced hourly water-level time series from 00{:}00 on 1 January 2023 to 00{:}00 on 1 January 2024. Based on the modeling results, we calculated the mean spring tidal range (MSR) and mean tidal range (MTR) for each observation point, and MSR was used as the primary tidal indicator in our analysis.

\subsubsection{Derived parameters}
Combining the beach profile, sediment, wave, and tidal datasets, we derived a suite of morphodynamic and hydrodynamic parameters for each beach. Morphological descriptors include beach slope, cross-shore width, and the presence/absence of the dry beach (i.e., the supratidal zone between the mean spring high water level and the foredune toe or vegetation, which remains subaerially exposed under normal hydrodynamic conditions). Sedimentological descriptors include sediment composition and grain-size parameters (e.g., mean grain size, sorting) derived from the sieve data \cite{folk1957brazos}. Hydrodynamic descriptors, such as the breaker wave height (Hb), the relative tidal range (RTR), and the dimensionless fall velocity ($\Omega$) linking wave and sediment properties, were calculated following published formulations \cite{wright1984morphodynamic}. Together, these variables provide a consistent set of environmental drivers for the MorphoGP framework and provide the basis for the category-specific equilibrium beach profile analysis.
\begin{figure*}[!ht]
\centering
\includegraphics[width=\textwidth]{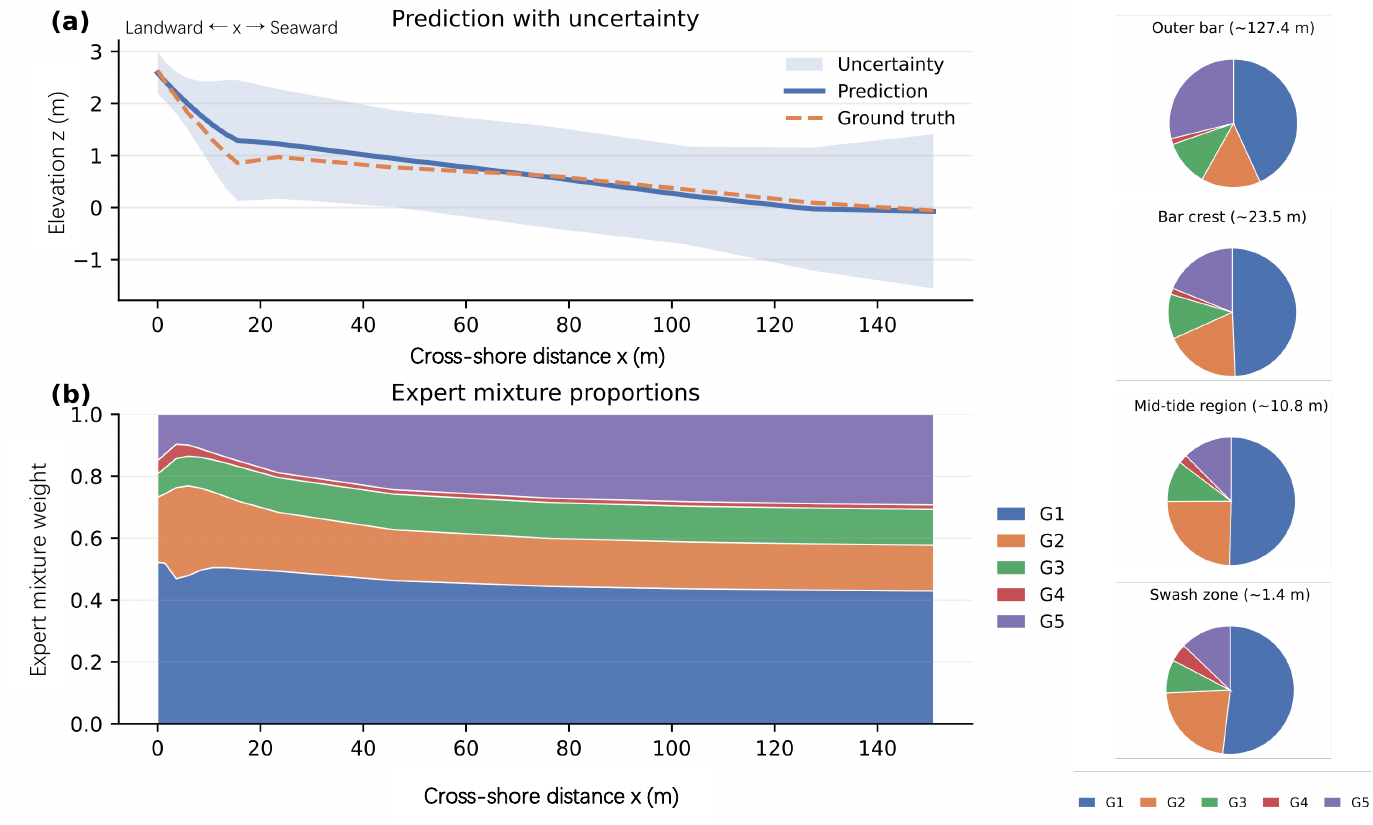}
\caption{\textbf{Expert decomposition of the predicted cross-shore profile.} \textbf{(a)} Predicted cross-shore profile with uncertainty. The solid curve shows the MorphoGP prediction, while the dashed curve denotes the observed elevation. The shaded band represents predictive uncertainty, providing an overall assessment of profile-scale accuracy. \textbf{(b)} Continuous expert mixture proportions along the profile. The stacked areas show the normalized mixture weights of individual experts, obtained by applying a softmax transformation to the pre-softmax gating logits, illustrating how expert responsibility varies smoothly in space. Local expert contributions at representative cross-shore locations. At selected positions along the transect, pie charts visualize the corresponding expert weight distributions, highlighting heterogeneous regions where different experts dominate the prediction.} \label{fig:expert_decomposition}
\end{figure*}
\subsection{Implementation Details}
All experiments were executed on a single NVIDIA A100 Tensor Core GPU. We adopted five-fold cross-validation, with splits performed at the beach level to avoid spatial leakage. Reported results are averaged over the five test folds.

For reproducibility, a fixed random seed was used for data splitting, parameter initialization, and mini-batch sampling. All models were trained under the same optimization and stopping protocol for a fair comparison. We used the Adam optimizer with a learning rate of \(1\times10^{-3}\). Training was run for up to 200 epochs with early stopping based on validation performance, and a batch size of 16 was used throughout.

\subsection{Performance}

To validate the effectiveness of the proposed framework, we conduct a comparative evaluation against a diverse set of baselines. All methods are assessed under the same evaluation protocol, and performance is reported using three standard regression metrics: the coefficient of determination (\(R^2\)), root mean square error (RMSE)~\cite{willmott2005advantages}, and mean absolute error (MAE).

We compare MorphoGP with representative data-driven regressors, including K-Nearest Neighbors (KNN)~\cite{cover1967nearest}, Linear Regression, a Multilayer Perceptron (MLP), Random Forest (RF)~\cite{breiman2001random}, Support Vector Regression (SVR)~\cite{cortes1995support}, and a Transformer-based model~\cite{vaswani2017attention}. In addition, two physics-inspired analytical baselines are implemented using classical one-dimensional equilibrium-profile formulations: (i) a Bruun/Dean-type profile \(y = A x^{2/3}\)~\cite{bruun1954,dean1991}, and (ii) an exponential profile \(y = B(1-e^{-kx})\)~\cite{larson1999}. For these analytical baselines, the shape parameters are not fitted separately for each profile. Instead, the coefficients \(A\), \(B\), and \(k\) are predicted from the same set of wave, tidal, and sediment descriptors used by the transformer-based models: the coefficients are estimated on the training set via regression and then used to generate predicted equilibrium profiles on the test set, from which the target morphodynamic indicator is derived.

As shown in Table~\ref{tab:performance}, MorphoGP achieves the highest \(R^2\) (\(0.942\)), indicating the strongest ability to explain the variance of the observations, while also attaining the lowest RMSE (\(0.297\)) and MAE (\(0.209\)). The improvement over the best-performing conventional data-driven baseline (RF, \(R^2 = 0.654\), RMSE = 0.730, MAE = 0.462) suggests that the proposed nonparametric probabilistic formulation enhances both predictive accuracy and generalization for macrotidal equilibrium-profile prediction. The analytical baselines provide a reference to earlier theoretical work by imposing classical profile shapes through \(y = A x^{2/3}\) and \(y = B(1-e^{-kx})\); their results further highlight the limitations of tide-insensitive parametric forms when applied to diverse macrotidal beaches where tidal processes strongly influence profile morphology.

\begin{table*}[h]
\renewcommand{\arraystretch}{1.25}
\caption{Ablation study on key components of MorphoGP. The \textbf{Reference} is the full configuration (ContourCluster + GP predictor + RBF-ARD kernel). Each variant modifies only the component indicated by its category while keeping all other components identical to the reference. The random grouping control keeps the same number of groups, GP experts, and gating structure as MorphoGP, but replaces ContourCluster labels with random group assignments. Results for random grouping are reported as mean \(\pm\) standard deviation over five random seeds. Boldface indicates the best result across all rows.}
\label{tab:ablation_all}
\centering
\footnotesize
\begin{tabular}{llccc}
\hline
\textbf{Category} & \textbf{Variant} &
\textbf{\(\bm{R^2}\) (\(\uparrow\))} & \textbf{RMSE (\(\downarrow\))} & \textbf{MAE (\(\downarrow\))} \\
\hline
Reference & \textbf{Full MorphoGP (ContourCluster + GP + RBF-ARD)} &
\textbf{0.942} & \textbf{0.297} & \textbf{0.209} \\
\hline
Classification & w/o ContourCluster (Single GP, RBF-ARD) &
0.664 & 0.718 & 0.463 \\
& Random grouping (GP experts + Gating Net) &
0.612 & 0.772 & 0.507 \\
\hline
Predictor & MLP (ContourCluster; replaces GP) &
0.876 & 0.436 & 0.299 \\
& Transformer (ContourCluster; replaces GP) &
0.920 & 0.350 & 0.257 \\
\hline
Kernel & Matérn-ARD (\(\nu = 3/2\)) (ContourCluster + GP) &
0.935 & 0.316 & 0.224 \\
& Rational Quadratic-ARD (ContourCluster + GP) &
0.930 & 0.321 & 0.232 \\
\hline
\end{tabular}
\end{table*}
Process-based numerical morphodynamic models, such as XBeach and Delft3D, are not included as direct baselines in this comparison. These models primarily target time-evolving simulations of coastal change under prescribed forcing and typically require detailed, site-specific boundary conditions, long-term hydrodynamic and sediment time series, and extensive calibration for each location. Such inputs are not consistently available for the large, multi-site dataset considered here. Moreover, MorphoGP is a reduced-order framework that predicts static morphodynamic indicators from aggregated environmental descriptors, whereas process-based models are designed to resolve transient processes. A direct benchmark would therefore be both computationally prohibitive and conceptually mismatched. We instead focus on baselines that operate on the same input feature space and are intended for fast, large-sample prediction, while reporting per-profile analytical fits separately as curve-fitting references.

Beyond aggregate error metrics, Fig.~\ref{fig:expert_decomposition} offers qualitative evidence that MorphoGP captures cross-shore morphological structure. Across representative transects, the predicted profiles reproduce the observed equilibrium shapes, while the 95\% predictive confidence intervals adapt to location-dependent variability along the profile, indicating where predictions are more or less reliable.

\subsection{Morphological Validity of ContourCluster}
To examine whether the learned ContourCluster categories represent meaningful profile-geometry regimes rather than arbitrary numerical partitions, we compare them with the classical Masselink--Short \(\Omega\)--RTR morphodynamic classification. The \(\Omega\)--RTR framework remains an important process-oriented classification for tide-influenced beaches because it links beach state to dimensionless fall velocity and relative tide range. However, it is not designed to directly optimize the geometric compactness of complete cross-shore profiles.

As shown in Table~\ref{tab:ms_contour_comparison}, the two classification schemes show low label agreement, with an NMI of 0.104. This indicates that ContourCluster does not simply reproduce the traditional \(\Omega\)--RTR classes. In the raw profile-geometry space, ContourCluster achieves a higher silhouette score than the Masselink--Short classification (0.191 versus -0.355), suggesting more compact and better separated profile groups. These results suggest that the Masselink--Short framework remains physically meaningful as a process-oriented morphodynamic classification, while ContourCluster provides a complementary morphology-driven partition that is more suitable for defining geometrically coherent profile regimes in the present dataset.

\begin{table}[t]
\centering
\footnotesize
\renewcommand{\arraystretch}{1.2}
\caption{Comparison between the Masselink--Short \(\Omega\)--RTR classification and ContourCluster. NMI measures the label agreement between the two schemes; the silhouette score evaluates compactness and separation in the raw profile-geometry space.}
\label{tab:ms_contour_comparison}
\begin{tabular}{lcc}
\hline
\textbf{Metric} & \textbf{Masselink--Short} & \textbf{ContourCluster} \\
\hline
NMI with the other scheme & 0.104 & 0.104 \\
Silhouette score & -0.355 & 0.191 \\
\hline
\end{tabular}
\end{table}

\begin{figure}[t]
\centering
\includegraphics[width=\linewidth]{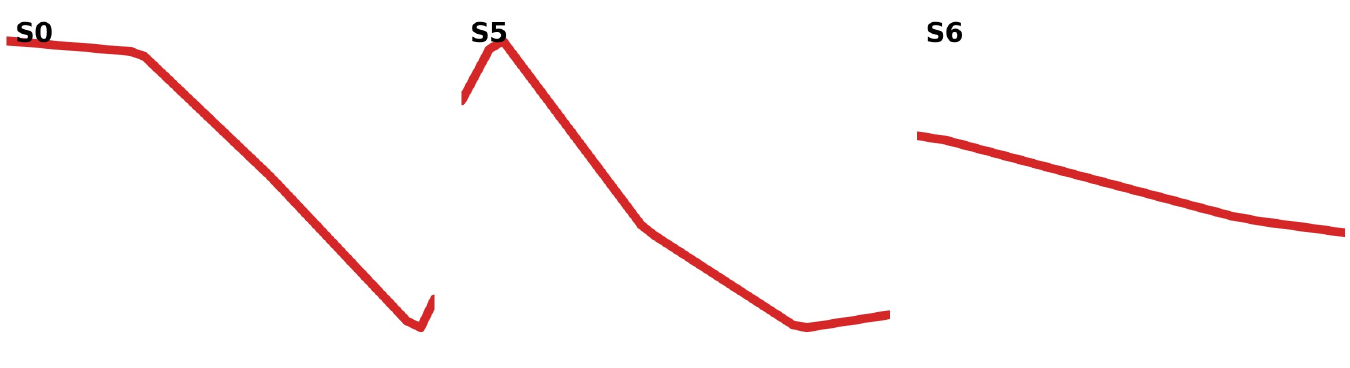}
\caption{Representative learned shapelets projected back to observed cross-shore profile segments. S0 shows a low-gradient concave transition, S5 shows a convex slope-break-like transition, and S6 shows a steep beachface-like segment. These interpretations describe local profile-geometry patterns associated with learned subsequences and should not be regarded as deterministic geomorphological labels or direct process-level explanations.}
\label{fig:shapelet_interpretation}
\end{figure}

We further inspect representative learned shapelets by projecting them back to the profile segments where they show high similarity. Fig.~\ref{fig:shapelet_interpretation} shows three representative local subsequences. These examples suggest that the learned shapelets are associated with recognizable local profile-geometry patterns: S0 corresponds to a low-gradient concave transition, S5 resembles a convex slope-break transition, and S6 captures a steep beachface-like segment with relatively stable slope.

These interpretations should be considered cautious and geometry-based. The learned shapelets are statistical subsequences extracted from observed profiles, and they should not be interpreted as deterministic geomorphological units or direct process-level representations. Nevertheless, their correspondence with local profile patterns helps reduce the black-box nature of ContourCluster and supports its use as a morphology-aware clustering module.

\subsection{Ablation and Sensitivity Analysis}

To investigate the contribution of each component in the proposed MorphoGP framework, we conducted a series of ablation experiments focusing on three aspects: (1) the contour-based morphodynamic classification module, (2) the choice of prediction architecture, and (3) the selection of covariance kernel within the Gaussian Process (GP) predictor. All experiments used the same data split and training protocol for a fair comparison. The results are summarized in Table~\ref{tab:ablation_all}.

Removing the \emph{ContourCluster} module and training a single GP expert across all beaches (``w/o ContourCluster'') leads to a clear drop in performance compared with the full MorphoGP configuration, with \(R^2\) decreasing from 0.942 to 0.664 and RMSE increasing from 0.297 to 0.718. This result indicates that a single global GP is insufficient to represent the heterogeneous forcing--profile relationships across different beach morphologies. To further examine whether the improvement comes from morphology-aware grouping rather than simply from using multiple experts, we added a random grouping control. In this variant, the number of groups, GP experts, and gating structure are kept unchanged, while the ContourCluster labels are replaced by random group assignments. Across five random seeds, the random grouping variant obtains \(R^2=0.612\), RMSE \(=0.772\), and MAE \(=0.507\), which are substantially worse than the full MorphoGP results. This suggests that the learned contour-based partition provides more coherent domains for category-specific regression than arbitrary grouping.

To assess the impact of the prediction architecture, we replaced the GP with a multilayer perceptron (MorphoGP-MLP) and a Transformer-based regressor (MorphoGP-Trans), while keeping the input features and the ContourCluster module unchanged. Both deep-learning variants achieve reasonably strong performance, but they remain consistently inferior to the GP-based MorphoGP in all metrics. The GP predictor attains the highest \(R^2\) and the lowest RMSE and MAE, suggesting that its probabilistic structure provides smoother regression and better uncertainty handling in data-sparse regions, which is crucial for morphodynamic prediction.

Finally, we examined the influence of the GP kernel by testing three widely used covariance functions: the Radial Basis Function (RBF), a Matérn kernel with \(\nu = 3/2\), and the Rational Quadratic kernel. These kernels span a spectrum from very smooth priors (RBF) to rougher local variability (Matérn) and multi--scale behavior (Rational Quadratic). As shown in Table~\ref{tab:ablation_all}, MorphoGP with the RBF kernel yields the best overall trade-off among \(R^2\), RMSE, and MAE, and is therefore adopted as the default kernel in the final model. The relatively small performance differences across kernels indicate that MorphoGP is robust to reasonable choices of kernel smoothness and length scale, rather than relying on a highly tuned, exotic kernel.

In addition to component ablations, we analyze the sensitivity of the ContourCluster module to the number of clusters \(k\), which controls the granularity of the morphodynamic partition. We sweep \(k \in \{2,3,\dots,8\}\) under identical beach-level cross-validation splits and compute the fold-averaged Silhouette coefficient for each \(k\). Across folds, the Silhouette score increases from small \(k\), peaks at \(k=5\), and then decreases for larger \(k\), indicating that overly coarse or overly fine partitions degrade clustering compactness and separation. To avoid selection bias, the optimal \(k^\star\) is selected using only the training split within each fold and then fixed when evaluating the corresponding test split. Under this protocol, the best average performance is achieved at \(k^\star=5\) with a fold-averaged Silhouette coefficient of \(0.4235\).

\subsection{Region-Held-Out Transfer and OOD Uncertainty Assessment}
\label{sec:ood_uncertainty}
\begin{table}[h]
\renewcommand{\arraystretch}{1.2}
\caption{Region-held-out OOD evaluation for MorphoGP. In each run, one region is withheld for testing and the model is trained on the remaining regions. Reported RMSE and mean predictive variance \(\overline{E}_{\mathrm{unc}}\) are computed over all test points in the withheld region; ``Overall'' denotes the random-split reference.}
\label{tab:ood_uncertainty_by_region}
\centering
\footnotesize
\begin{tabular}{lcc}
\hline
\textbf{Evaluation setting} & \textbf{RMSE (\(\downarrow\))} & \textbf{\(\bm{\overline{E}}_{\mathrm{unc}}\) (\(\uparrow\))}
 \\
\hline
Overall (random split / test folds) & 0.297 & 0.071 \\
\hline
Hainan held out    & 0.328 & 0.105 \\
Fujian held out    & 0.351 & 0.113 \\
Guangdong held out & 0.314 & 0.097 \\
\hline
\end{tabular}
\end{table}
Global error metrics such as RMSE and \(R^2\) provide an overall indication of predictive skill, but they do not reveal whether the model can recognize situations in which its predictions are less reliable. This is particularly important when transferring equilibrium beach-profile models to new coastal settings where forcing--morphology combinations may differ from those represented during training. We therefore assess regional transferability and uncertainty response using a region-held-out evaluation protocol and test whether MorphoGP assigns higher predictive uncertainty under geographic domain shift. It should be noted that this setting represents a moderate regional domain shift within the South China coastal dataset, rather than a strict extrapolation to entirely different coastal systems.
\begin{figure*}[!h]
    \centering
    \includegraphics[width=\textwidth]{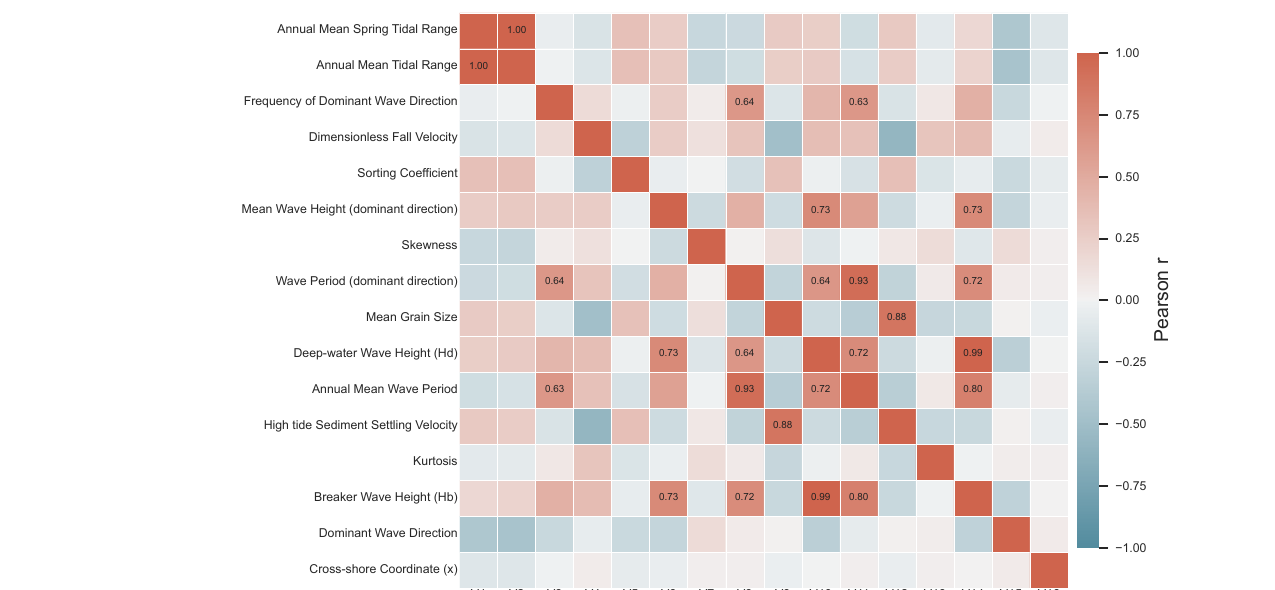}
    \caption{Pearson correlation heatmap of the 16 selected input variables. The analysis was conducted using all profile-point samples. Strong correlations are mainly observed within tidal-range, wave-height, and wave-period variable groups, indicating potential redundancy that should be considered when interpreting ARD-based feature relevance.}
    \label{fig:feature_corr}
\end{figure*}
For any forcing vector \(c\) (wave, tide, sediment, and associated descriptors) and cross-shore coordinate \(x\), MorphoGP yields a predictive distribution \(p(y \mid x,c)\) with mean \(\hat{y}(x,c)\) and variance \(\mathrm{Var}[\hat{y}(x,c)]\) (see (24)). We use the predictive variance as an uncertainty score,
\begin{equation}
E_{\mathrm{unc}}(x,c) = \mathrm{Var}\!\left[\hat{y}(x,c)\right],
\end{equation}
where larger values indicate lower model confidence. For reporting at the dataset/region level, we summarize uncertainty by the mean predictive variance \(\overline{E}_{\mathrm{unc}}=\frac{1}{N}\sum_{i=1}^{N} E_{\mathrm{unc},i}\) over all test points.

To mimic external validation, we perform leave-one-region-out evaluation over the three geographic groups with sufficient samples (Hainan, Fujian, and Guangdong). In each run, all profiles from one region are withheld as a test set and the model is trained using only the remaining regions. Profiles from other provinces (e.g., Zhejiang and Shandong) are included in the dataset but are not treated as a separate held-out group due to the limited number of samples. All preprocessing steps are fitted on the training regions only and then applied to the held-out region to prevent information leakage.

Table~\ref{tab:ood_uncertainty_by_region} reports RMSE and \(\overline{E}_{\mathrm{unc}}\) for each held-out region, together with the overall random-split evaluation for reference. Compared with the overall setting, all region-held-out runs exhibit higher \(\overline{E}_{\mathrm{unc}}\), suggesting that MorphoGP expresses reduced confidence when applied to unseen geographic regions. At the same time, the RMSE values remain relatively low, increasing only from 0.297 in the random-split setting to 0.314--0.351 in the region-held-out tests. This indicates that the three held-out regions are not completely dissimilar from the training regions in the feature and morphology spaces.

The relatively stable transfer performance can be explained by two factors. First, although Hainan, Fujian, and Guangdong are geographically distinct, they belong to the broader South China coastal setting and may share overlapping ranges of wave, tide, sediment, and morphological descriptors. Second, MorphoGP does not rely on geographic labels directly; instead, it organizes profiles according to learned morphological regimes and then applies category-specific Gaussian process experts. Therefore, if a held-out region contains beach-profile types and forcing combinations that are already represented in the training regions, the model can transfer reasonably well despite the geographic split.

However, this result should not be interpreted as evidence of universal OOD generalization. The current region-held-out protocol is limited by the available spatial coverage of the dataset and cannot fully evaluate model behavior under stronger domain shifts, such as regions with tidal ranges outside the training distribution, substantially different sediment supply, reef-controlled or engineered coastlines, storm-dominated disequilibrium profiles, or beach types absent from the learned ContourCluster categories. Under such conditions, both the predicted mean profiles and the uncertainty estimates may become less reliable. Therefore, the present evaluation demonstrates promising transferability under moderate regional shifts, while broader validation using more geographically and morphodynamically diverse datasets remains necessary for operational deployment.

\subsection{Relevance Analysis}
\label{sec:relevance_analysis}

\begin{figure*}[!h]
    \centering
    \includegraphics[width=\textwidth]{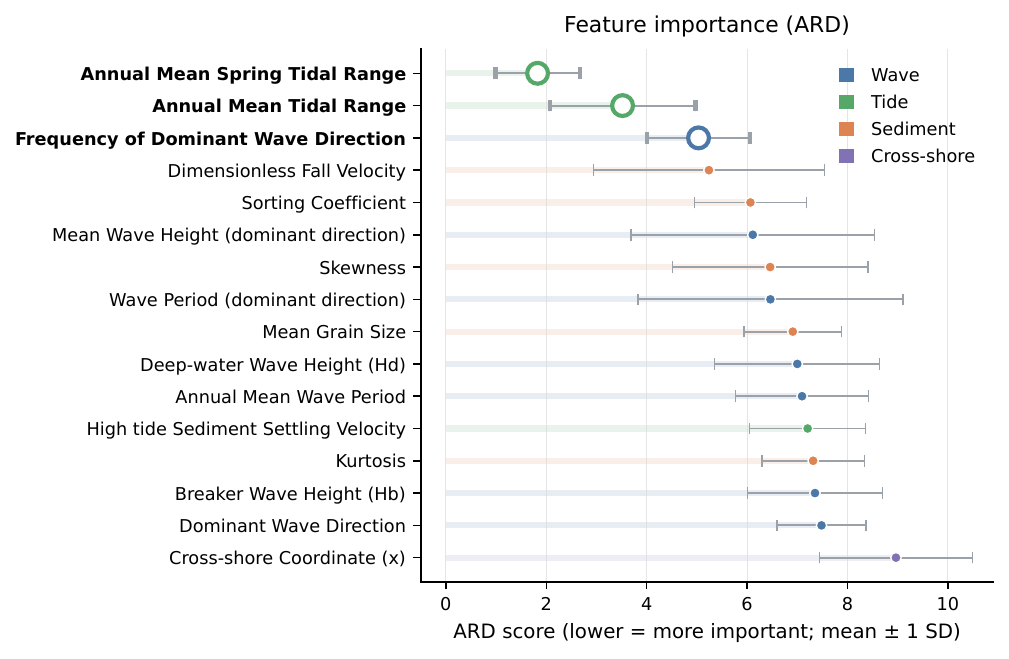}
    \caption{Feature-wise mean ARD values with variance for all input parameters in the MorphoGP framework. Smaller ARD values indicate higher feature relevance. The most relevant variables are mainly associated with tidal range, directional wave climate, and sediment/morphological descriptors. Because several input variables are strongly correlated, the ARD results are interpreted at the level of physically meaningful variable groups rather than as strict causal evidence for isolated variables.}
    \label{ARD}
\end{figure*}

To further interpret the internal decision-making process of the proposed MorphoGP model and to identify the dominant predictive descriptors of equilibrium beach morphology, we conducted a feature relevance analysis based on the Automatic Relevance Determination (ARD) mechanism within each Gaussian Process expert.

In this analysis, each input variable is assigned an independent characteristic length scale, which is optimized jointly with other GP hyperparameters during training. We report the learned ARD length scales, where smaller values indicate higher sensitivity of the model output to the corresponding input dimension.

To support interpretation, we examine pairwise linear correlations among the 16 input variables using Pearson correlation analysis (Fig.~\ref{fig:feature_corr}).

The correlation analysis reveals clear intra-group dependencies among several physically related variables. Strong correlations are observed within wave- and sediment-related descriptors, including Breaker Wave Height and Deepwater Wave Height (\(r=0.993\)), Wave Period and Annual Mean Wave Period (\(r=0.935\)), and Mean Grain Size and High-tide Sediment Settling Velocity (\(r=0.879\)). These results indicate that variables within the same physical category tend to share overlapping information and may partially represent similar underlying processes.

In contrast, correlations across different physical groups are generally weak. Tidal-range-related variables exhibit only weak linear relationships with wave- and sediment-related descriptors, with most correlation coefficients typically below \(|r| < 0.30\). This suggests limited linear redundancy between tidal forcing and other environmental descriptors in the dataset, and indicates that different physical drivers provide largely independent information.

These results suggest that the high ARD relevance of tidal-range-related variables is unlikely to be solely driven by linear collinearity with wave or sediment descriptors. Instead, tidal variables appear to provide complementary predictive information within the overall environmental descriptor space. The Pearson correlation analysis is used here solely to assess redundancy and dependence structure among input variables, thereby providing context for interpreting the ARD-based importance ranking.

Fig.~\ref{ARD} illustrates the feature-wise mean ARD values with standard deviation across 50 independent training runs, sorted in ascending order to emphasize the most influential features. As shown in the figure, Annual Mean Spring Tidal Range has the smallest ARD value, followed by Annual Mean Tidal Range and Frequency of Dominant Wave Direction, indicating that these variables show the highest sensitivity within the trained MorphoGP model. Combined with the Pearson correlation analysis, this result suggests that tidal-range-related descriptors provide strong and relatively independent predictive information for equilibrium beach-profile prediction, rather than simply acting as linear substitutes for wave- or sediment-related descriptors.

From the perspective of the main driving factors, the prediction of equilibrium beach profiles (EBP) for macrotidal beaches is broadly consistent with geomorphological expectations. Tidal range determines the extent of the intertidal zone and the frequency with which different parts of the beach are exposed to hydrodynamic forcing, thereby constraining the spatiotemporal scales of coastal processes acting on the profile. The presence or absence of a dry beach further reflects whether the equilibrium profile is in a self-stable state or in a stressed state under sediment deficit. Beach slope and cross-shore width collectively modulate wave-energy dissipation and the associated sediment redistribution across the intertidal zone, which are key processes governing the evolution of equilibrium profiles. However, these interpretations should be understood as physically plausible explanations of model sensitivity and predictive relevance, not as definitive causal attribution derived solely from ARD.

A second tier of variables, including Frequency of Dominant Wave Direction, the Dimensionless Fall Velocity, Dominant Wave Direction, and the Sorting Coefficient, also shows relatively low ARD values, suggesting appreciable but more nuanced roles in shaping the equilibrium state. These parameters primarily characterize the dominant wave climate and sediment properties, which adjust the detailed profile shape around the broader hydrodynamic and morphological template. Their relevance indicates that wave and sediment descriptors still provide useful predictive information, although their individual ARD rankings may be affected by correlations within each descriptor group.

In contrast, variables such as the cross-shore coordinate (\(x\)), Relative Tidal Range (RTR), Mean Grain Size (mm), deepwater wave height (\(H_d\)), and breaker wave height (\(H_b\)) appear toward the upper end of the ARD spectrum, indicating weaker direct sensitivity in the predictive model. Nevertheless, these variables provide important contextual information about spatial geometry, offshore hydrodynamics, and sediment characteristics, and they may still influence morphodynamic responses indirectly through nonlinear feature interactions within the mixture-of-experts architecture. In particular, the lower ARD relevance of \(H_d\) and \(H_b\) should be interpreted cautiously because these two variables are highly correlated with each other, and their predictive information may be partly represented by other wave-related descriptors.

The variance of ARD values across training runs provides further insight into the stability of feature influence. Tidal and morphological parameters such as Annual Mean Spring Tidal Range and Beach Slope exhibit relatively low variance, implying a consistent and robust contribution to the model predictions under different random initializations and resampling. By contrast, several wave-related descriptors, such as Mean Wave Height and Wave Period, display higher variance in their ARD values, likely reflecting interdependence with other hydrodynamic or sedimentary factors and greater sensitivity to sampling variability.

Overall, the ARD-based analysis suggests that tidal-range-related and morphological descriptors provide strong and stable predictive information for macrotidal equilibrium beach-profile prediction in MorphoGP. The Pearson correlation analysis further supports the reliability of this interpretation by showing that tidal descriptors are not highly linearly correlated with most wave- and sediment-related descriptors. Therefore, the high ARD relevance of tidal variables is unlikely to be caused solely by cross-category multicollinearity. This model-based finding is consistent with physical expectations for macrotidal environments, where strong tides and the associated morphological configuration influence nearshore hydrodynamics and sediment transport in ways that are more complex than in predominantly wave-dominated coasts.

\section{Conclusion}
\label{conclusion}

This study proposes MorphoGP, a category-specific Gaussian Process framework for equilibrium beach-profile prediction under varying tidal conditions. The framework integrates unsupervised contour-based classification, probabilistic GP regression, and a gating mechanism for expert aggregation, enabling morphology-conditioned prediction from environmental descriptors.

MorphoGP organizes beach profiles into data-driven morphological regimes and performs probabilistic prediction of equilibrium-like intertidal profiles from wave, tide, and sediment descriptors. Rather than explicitly simulating full coastal morphodynamic processes, it learns statistical relationships between environmental forcing and observed profile geometry.

Experiments on 183 beaches along the Chinese coast demonstrate that MorphoGP consistently outperforms conventional machine learning and deep learning baselines in terms of $R^2$, RMSE, and MAE. The ContourCluster module yields compact and well-separated morphological groups, and ablation studies confirm the effectiveness of both the clustering and Gaussian process expert components.

In addition, MorphoGP provides structured uncertainty estimates, where predictive variance increases under region-held-out evaluation, indicating reduced confidence under moderate domain shift. Feature relevance analysis further suggests that tidal-range and morphological descriptors play a dominant role in explaining variability in equilibrium profile shapes, while wave and sediment variables contribute complementary information.

The learned shapelets and expert representations capture recurring local geometric patterns in cross-shore profiles, providing an interpretable view of morphological variability at the sub-profile scale. However, these representations should be interpreted as statistical descriptors of observed geometry rather than direct process-based physical units.

Overall, MorphoGP provides a probabilistic and morphology-aware framework for equilibrium beach-profile prediction with quantified uncertainty. Its main contributions lie in morphology-driven regime decomposition, GP-based predictive modeling, and spatially varying expert aggregation. 

Future work will focus on extending the framework to fully continuous spatio-temporal profile evolution, improving physical constraints in the predictive model, and incorporating broader cross-regional datasets to enhance generalization under stronger domain shifts.

\section*{Acknowledgment}
The computations in this research were performed using the CFFF platform of Fudan University.

\appendices

\section{Per-Profile Fitted Analytical References}
\label{app:analytical_fitting}

\begin{table}[!h]
\centering
\caption{Per-profile least-squares fitted analytical references. These results are reported as oracle-style curve-fitting references rather than predictive baselines.}
\label{tab:appendix_analytical_fitting}
\begin{tabular}{lccc}
\toprule
\textbf{Model} & \textbf{\(R^2\)} & \textbf{RMSE (m)} & \textbf{MAE (m)} \\
\midrule
Bruun/Dean-type model & 0.9855 & 0.1494 & 0.1040 \\
Exponential model & 0.9963 & 0.0750 & 0.0473 \\
\bottomrule
\end{tabular}
\end{table}

To further clarify the distinction between profile prediction and post-hoc curve fitting, we provide additional per-profile least-squares fitting results for the classical analytical formulations. Unlike MorphoGP, which predicts unseen equilibrium beach profiles from environmental descriptors without using the target profile elevations during inference, the per-profile fitted analytical models use each observed target profile to estimate their parameters. Therefore, these results are reported as oracle-style curve-fitting references rather than strict predictive baselines.

For each observed test profile, the Bruun/Dean-type parameter \(A\) and the exponential parameters \(B\) and \(k\) were optimized independently by least squares. The fitted profiles were then evaluated using the same metrics as in the main experiments. As shown in Table~\ref{tab:appendix_analytical_fitting}, per-profile fitting substantially improves the analytical formulations, indicating that these parametric models have strong shape-fitting capacity when the target profile is available. However, this setting differs from the predictive task considered in the main comparison, where the target profile is unavailable at inference time.

\section{Definition of Input Parameters}
\begin{table*}[!h]

\centering

\caption{Definitions of the input parameters used in this study.}

\label{tab:appendix_input_parameters}

\small

\setlength{\tabcolsep}{4pt}

\renewcommand{\arraystretch}{1.12}

\begin{tabular}{p{3.2cm} p{1.8cm} p{1.5cm} p{9.2cm}}

\toprule

\textbf{Input parameter} & \textbf{Symbol} & \textbf{Unit} & \textbf{Definition / calculation} \\

\midrule

Cross-shore coordinate & \(x\) & m & Horizontal coordinate perpendicular to the shoreline. \\

Mean spring tidal range & MSR & m & Vertical difference between mean high water springs and mean low water springs. \\

Mean tidal range & MTR & m & Average vertical difference between adjacent high and low tides. \\

Relative tidal range & RTR & -- & \(\mathrm{RTR}=\mathrm{MSR}/H_b\). \\

Breaker wave height & \(H_b\) & m & \(H_b=0.39g^{1/5}(TH_d^2)^{2/5}\). \\

Deep-water wave height & \(H_d\) & m & Wave height in deep-water conditions. \\

Wave period & \(T\) & s & Time interval between two consecutive wave crests or troughs. \\

Dominant wave direction & \(\theta\) & \(^{\circ}\) & Wave direction with the highest integrated energy. \\

Frequency of dominant wave direction & -- & \% & Occurrence frequency of waves from the dominant direction. \\

Dimensionless fall velocity & \(\Omega\) & -- & \(\Omega=H_b/(\omega_s T)\). \\

High-tide sediment fall velocity & \(\omega_s\) & m s\(^{-1}\) & Sediment fall velocity at high-tide level. \\

Mean grain size & \(M_z/d_m\) & \(\Phi\)/mm & Average sediment particle size. \\

Sorting coefficient & \(\delta\) & -- & \(\delta=(\Phi_{84}-\Phi_{16})/4+(\Phi_{95}-\Phi_{5})/6.6\). \\

Skewness & Sk & -- & Grain-size distribution asymmetry. \\

Kurtosis & \(K\) & -- & Peakedness of the grain-size distribution. \\

Beach slope & \(\tan\beta\) & -- & Intertidal beach slope derived from the surveyed profile. \\

\bottomrule

\end{tabular}

\vspace{0.5em}

\begin{minipage}{0.96\textwidth}

\footnotesize

\textit{Note:} \(\Phi_x\) denotes the grain-size value at which the cumulative percentage reaches \(x\%\). Due to field-survey limitations, the lowest surveyed elevation may not always reach the mean low water spring level.

\end{minipage}

\end{table*}
\label{app:input_parameters}

Table~\ref{tab:appendix_input_parameters} summarizes the input parameters used in this study. Detailed derivations of standard morphodynamic parameters, such as \(\Omega\) and RTR, follow conventional definitions.

\ifCLASSOPTIONcaptionsoff
  \newpage
\fi

\bibliographystyle{IEEEtran}
\bibliography{bibtex/bib/Shore_Processes_and_Shoreline_Development}

\begin{thebibliography}{10}
\providecommand{\url}[1]{#1}
\csname url@samestyle\endcsname
\providecommand{\newblock}{\relax}
\providecommand{\bibinfo}[2]{#2}
\providecommand{\BIBentrySTDinterwordspacing}{\spaceskip=0pt\relax}
\providecommand{\BIBentryALTinterwordstretchfactor}{4}
\providecommand{\BIBentryALTinterwordspacing}{\spaceskip=\fontdimen2\font plus
\BIBentryALTinterwordstretchfactor\fontdimen3\font minus \fontdimen4\font\relax}
\providecommand{\BIBforeignlanguage}[2]{{%
\expandafter\ifx\csname l@#1\endcsname\relax
\typeout{** WARNING: IEEEtran.bst: No hyphenation pattern has been}%
\typeout{** loaded for the language `#1'. Using the pattern for}%
\typeout{** the default language instead.}%
\else
\language=\csname l@#1\endcsname
\fi
#2}}
\providecommand{\BIBdecl}{\relax}
\BIBdecl

\bibitem{dean1991}
R.~G. Dean, ``Equilibrium beach profiles: Characteristics and applications,'' \emph{Journal of Coastal Research}, vol.~7, no.~1, pp. 53--84, 1991.

\bibitem{johnson1919}
D.~W. Johnson, \emph{Shore Processes and Shoreline Development}.\hskip 1em plus 0.5em minus 0.4em\relax New York: Prentice Hall, 1919, p. 584.

\bibitem{cornaglia1889}
P.~Cornaglia, ``Delle spiaggie,'' \emph{Atti della Classe di Scienze Fisiche, Matematiche e Naturali}, vol.~5, pp. 284--304, 1889.

\bibitem{short1991macro}
A.~Short, ``Macro-meso tidal beach morphodynamics: an overview,'' \emph{Journal of Coastal Research}, pp. 417--436, 1991.

\bibitem{wright1982morphodynamics}
L.~Wright, P.~Nielsen, A.~Short, and M.~Green, ``Morphodynamics of a macrotidal beach,'' \emph{Marine geology}, vol.~50, no. 1-2, pp. 97--127, 1982.

\bibitem{bruun1954}
P.~Bruun, \emph{Coast erosion and the development of beach profiles}.\hskip 1em plus 0.5em minus 0.4em\relax US Beach Erosion Board, 1954, vol.~44.

\bibitem{dean1977}
R.~G. Dean, \emph{Equilibrium beach profiles: US Atlantic and Gulf coasts}.\hskip 1em plus 0.5em minus 0.4em\relax Department of Civil Engineering, University of Delaware Newark, Delaware, 1977, vol.~12.

\bibitem{bodge1992}
K.~R. Bodge, ``Representing equilibrium beach profiles with an exponential expression,'' \emph{Journal of Coastal Research}, vol.~8, no.~1, pp. 47--55, 1992.

\bibitem{lee1994}
P.~Z.-F. Lee, ``The submarine equilibrium profile: A physical model,'' \emph{Journal of Coastal Research}, vol.~10, no.~1, pp. 1--17, 1994.

\bibitem{larson1999}
M.~Larson, N.~C. Kraus, and R.~A. Wise, ``Equilibrium beach profiles under breaking and non-breaking waves,'' \emph{Coastal Engineering}, vol.~36, no.~1, pp. 59--85, 1999.

\bibitem{stokes2015observation}
C.~Stokes, M.~Davidson, and P.~Russell, ``Observation and prediction of three-dimensional morphology at a high-energy macrotidal beach,'' \emph{Geomorphology}, vol. 243, pp. 1--13, 2015.

\bibitem{wei2025analysis}
Y.~Wei, H.~Qi, G.~Liu, F.~Cai, Y.~He, K.~Hua, and S.~Zhao, ``Analysis of the morphodynamic characteristics of macrotidal beaches on the western coast of the taiwan strait,'' \emph{Regional Studies in Marine Science}, p. 104429, 2025.

\bibitem{anthony2004morphodynamics}
E.~Anthony, F.~Levoy, and O.~Monfort, ``Morphodynamics of intertidal bars on a megatidal beach, merlimont, northern france,'' \emph{Marine geology}, vol. 208, no.~1, pp. 73--100, 2004.

\bibitem{hashemi2010}
M.~Hashemi, Z.~Ghadampour, and S.~Neill, ``Using an artificial neural network to model seasonal changes in beach profiles,'' \emph{Ocean Engineering}, vol.~37, no. 14-15, pp. 1345--1356, 2010.

\bibitem{de2024}
W.~W. de~Melo, J.~Pinho, and I.~Iglesias, ``A data model to forecast the morphological evolution of multiple beach profiles,'' \emph{Coastal Engineering}, vol. 192, p. 104574, 2024.

\bibitem{khan2024}
A.~R. Khan, M.~S.~B. Ab~Razak, B.~B. Yusuf, H.~Z. B.~M. Shafri, and N.~B. Mohamad, ``Future prediction of coastal recession using convolutional neural network,'' \emph{Estuarine, Coastal and Shelf Science}, vol. 299, p. 108667, 2024.

\bibitem{beuzen2019}
T.~Beuzen, E.~B. Goldstein, and K.~D. Splinter, ``Ensemble models from machine learning: an example of wave runup and coastal dune erosion,'' \emph{Natural Hazards and Earth System Sciences}, vol.~19, no.~10, pp. 2295--2309, 2019.

\bibitem{adusumilli2024}
S.~Adusumilli, N.~Cirrito, L.~Engeman, J.~W. Fiedler, R.~Guza, A.~M. Lange, M.~A. Merrifield, W.~O'Reilly, and A.~P. Young, ``Predicting shoreline changes along the california coast using deep learning applied to satellite observations,'' \emph{Journal of Geophysical Research: Machine Learning and Computation}, vol.~1, no.~3, p. e2024JH000172, 2024.

\bibitem{demelo2024}
W.~W. de~Melo, J.~Pinho, and I.~Iglesias, ``A data model to forecast the morphological evolution of multiple beach profiles,'' \emph{Coastal Engineering}, vol. 192, p. 104574, 2024.

\bibitem{muroi2025}
S.~K. Muroi, E.~Bertone, N.~Cartwright, and F.~Alvarez, ``Machine learning methods for predicting shoreline change from submerged breakwater simulations,'' \emph{Engineering Applications of Artificial Intelligence}, vol. 152, p. 110726, 2025.

\bibitem{anthony2002between}
E.~J. Anthony and J.~D. Orford, ``Between wave-and tide-dominated coasts: the middle ground revisited,'' \emph{Journal of Coastal Research}, no.~36, pp. 8--15, 2002.

\bibitem{karunarathna2016linkages}
H.~Karunarathna, J.~Horrillo-Caraballo, Y.~Kuriyama, H.~Mase, R.~Ranasinghe, and D.~E. Reeve, ``Linkages between sediment composition, wave climate and beach profile variability at multiple timescales,'' \emph{Marine Geology}, vol. 381, pp. 194--208, 2016.

\bibitem{zhang2023classification}
L.~Zhang, H.~Xing, X.~Zhang, H.~Li, Z.~Liu, H.~Shi, and Z.~You, ``Classification of beach dynamic geomorphic features in the shandong peninsula,'' \emph{Mar Sci}, vol.~47, no.~02, pp. 10--19, 2023.

\bibitem{ding2020beach}
Y.~Ding, J.~Yu, and H.~Cheng, ``Beach morphodynamic characteristics and classifications on the straight coastal sectors in the west guangdong,'' \emph{Journal of Geographical Sciences}, vol.~30, no.~7, pp. 1179--1194, 2020.

\bibitem{lesser2004development}
G.~R. Lesser, J.~v. Roelvink, J.~T.~M. van Kester, and G.~Stelling, ``Development and validation of a three-dimensional morphological model,'' \emph{Coastal engineering}, vol.~51, no. 8-9, pp. 883--915, 2004.

\bibitem{warner2008development}
J.~C. Warner, C.~R. Sherwood, R.~P. Signell, C.~K. Harris, and H.~G. Arango, ``Development of a three-dimensional, regional, coupled wave, current, and sediment-transport model,'' \emph{Computers \& geosciences}, vol.~34, no.~10, pp. 1284--1306, 2008.

\bibitem{haidvogel2008ocean}
D.~B. Haidvogel, H.~Arango, W.~P. Budgell, B.~D. Cornuelle, E.~Curchitser, E.~Di~Lorenzo, K.~Fennel, W.~R. Geyer, A.~J. Hermann, L.~Lanerolle \emph{et~al.}, ``Ocean forecasting in terrain-following coordinates: Formulation and skill assessment of the regional ocean modeling system,'' \emph{Journal of computational physics}, vol. 227, no.~7, pp. 3595--3624, 2008.

\bibitem{roelvink2010xbeach}
D.~Roelvink, A.~Reniers, A.~Van~Dongeren, J.~Van Thiel~de Vries, J.~Lescinski, and R.~McCall, ``Xbeach model description and manual,'' \emph{Unesco-IHE Institute for Water Education, Deltares and Delft University of Tecnhology. Report June}, vol.~21, no. 2010, p.~2, 2010.

\bibitem{du2010modelling}
Y.~Du, S.~Pan, and Y.~Chen, ``Modelling the effect of wave overtopping on nearshore hydrodynamics and morphodynamics around shore-parallel breakwaters,'' \emph{Coastal Engineering}, vol.~57, no.~9, pp. 812--826, 2010.

\bibitem{kroon2025parameter}
A.~Kroon, J.~C. Christiaanse, A.~P. Luijendijk, M.~A.~d. Schipper, and R.~Ranasinghe, ``Parameter uncertainty in medium-term coastal morphodynamic modeling,'' \emph{Scientific Reports}, vol.~15, no.~1, p. 18471, 2025.

\bibitem{hsu1986two}
T.-W. Hsu, S.-R. Liaw, S.-K. Wang, and S.-H. Ou, ``Two-dimensional empirical eigenfunction model for the analysis and prediction of beach profile changes,'' in \emph{Coastal Engineering 1986}, 1986, pp. 1180--1195.

\bibitem{disdier2025predicting}
E.~Disdier, R.~Almar, R.~Benshila, M.~Al~Najar, R.~Chassagne, D.~Mukherjee, and D.~G. Wilson, ``Predicting beach profiles with machine learning from offshore wave reflection spectra,'' \emph{Environmental Modelling \& Software}, vol. 183, p. 106221, 2025.

\bibitem{kim2025prediction}
J.~Kim, T.~Kim, S.~Chang, J.~Kim, and I.~Kim, ``Prediction of beach profile changes using spatiotemporal graph neural networks for beach morphological evolution modeling,'' \emph{Ocean Engineering}, vol. 340, p. 122282, 2025.

\bibitem{williams2006gaussian}
C.~K. Williams and C.~E. Rasmussen, \emph{Gaussian processes for machine learning}.\hskip 1em plus 0.5em minus 0.4em\relax MIT press Cambridge, MA, 2006, vol.~2, no.~3.

\bibitem{masselink1993effect}
G.~Masselink and A.~D. Short, ``The effect of tide range on beach morphodynamics and morphology: a conceptual beach model,'' \emph{Journal of coastal research}, pp. 785--800, 1993.

\bibitem{ye2009shapelets}
L.~Ye and E.~Keogh, ``Time series shapelets: A new primitive for data mining,'' in \emph{Proceedings of the 15th ACM SIGKDD International Conference on Knowledge Discovery and Data Mining}, 2009, pp. 947--956.

\bibitem{lines2012shapelet}
J.~Lines, L.~M. Davis, J.~Hills, and A.~Bagnall, ``A shapelet transform for time series classification,'' in \emph{Proceedings of the 18th ACM SIGKDD International Conference on Knowledge Discovery and Data Mining}, 2012, pp. 289--297.

\bibitem{hills2014classification}
J.~Hills, J.~Lines, E.~Baranauskas, J.~Mapp, and A.~Bagnall, ``Classification of time series by shapelet transformation,'' \emph{Data Mining and Knowledge Discovery}, vol.~28, no.~4, pp. 851--881, 2014.

\bibitem{grabocka2014learning}
J.~Grabocka, N.~Schilling, M.~Wistuba, and L.~Schmidt-Thieme, ``Learning time-series shapelets,'' in \emph{Proceedings of the 20th ACM SIGKDD international conference on Knowledge discovery and data mining}, 2014, pp. 392--401.

\bibitem{riazi2021subaerial}
A.~Riazi and P.~A. Slovinsky, ``Subaerial beach profiles classification: An unsupervised deep learning approach,'' \emph{Continental Shelf Research}, vol. 226, p. 104508, 2021.

\bibitem{toms2003piecewise}
J.~Toms and M.~Lesperance, ``Piecewise regression: A tool for identifying ecological thresholds,'' \emph{Ecology}, vol.~84, no.~8, pp. 2034--2041, 2003.

\bibitem{hamilton1989new}
J.~Hamilton, ``A new approach to the economic analysis of nonstationary time series and the business cycle,'' \emph{Econometrica}, vol.~57, no.~2, pp. 357--384, 1989.

\bibitem{dietterich2000ensemble}
T.~Dietterich, ``Ensemble methods in machine learning,'' in \emph{Multiple Classifier Systems}.\hskip 1em plus 0.5em minus 0.4em\relax Springer, 2000, pp. 1--15.

\bibitem{rakthanmanon2013addressing}
T.~Rakthanmanon, B.~Campana, A.~Mueen, G.~Batista, B.~Westover, Q.~Zhu, J.~Zakaria, and E.~Keogh, ``Addressing big data time series: Mining trillions of time series subsequences under dynamic time warping,'' \emph{ACM Transactions on Knowledge Discovery from Data (TKDD)}, vol.~7, no.~3, pp. 1--31, 2013.

\bibitem{vaswani2017attention}
A.~Vaswani, N.~Shazeer, N.~Parmar, J.~Uszkoreit, L.~Jones, A.~N. Gomez, {\L}.~Kaiser, and I.~Polosukhin, ``Attention is all you need,'' \emph{Advances in neural information processing systems}, vol.~30, 2017.

\bibitem{hermans2017defense}
A.~Hermans, L.~Beyer, and B.~Leibe, ``In defense of the triplet loss for person re-identification,'' \emph{arXiv preprint arXiv:1703.07737}, 2017.

\bibitem{mackay1992bayesian}
D.~J. MacKay, ``Bayesian interpolation,'' \emph{Neural computation}, vol.~4, no.~3, pp. 415--447, 1992.

\bibitem{folk1957brazos}
R.~L. Folk and W.~C. Ward, ``Brazos river bar [texas]; a study in the significance of grain size parameters,'' \emph{Journal of sedimentary research}, vol.~27, no.~1, pp. 3--26, 1957.

\bibitem{benoit1996tomawac}
M.~Benoit, F.~Marcos, and F.~Becq, ``Development of a third generation shallow-water wave model with unstructured spatial meshing,'' in \emph{Proceedings of the 25th International Conference on Coastal Engineering (ICCE)}, Orlando, Florida, USA, 1996, pp. 465--478.

\bibitem{wright1984morphodynamic}
L.~D. Wright and A.~D. Short, ``Morphodynamic variability of surf zones and beaches: a synthesis,'' \emph{Marine geology}, vol.~56, no. 1-4, pp. 93--118, 1984.

\bibitem{willmott2005advantages}
C.~J. Willmott and K.~Matsuura, ``Advantages of the mean absolute error (mae) over the root mean square error (rmse) in assessing average model performance,'' \emph{Climate research}, vol.~30, no.~1, pp. 79--82, 2005.

\bibitem{cover1967nearest}
T.~Cover and P.~Hart, ``Nearest neighbor pattern classification,'' \emph{IEEE transactions on information theory}, vol.~13, no.~1, pp. 21--27, 1967.

\bibitem{breiman2001random}
L.~Breiman, ``Random forests,'' \emph{Machine learning}, vol.~45, no.~1, pp. 5--32, 2001.

\bibitem{cortes1995support}
C.~Cortes and V.~Vapnik, ``Support-vector networks,'' \emph{Machine learning}, vol.~20, no.~3, pp. 273--297, 1995.

\end{thebibliography}

\vspace{11pt}
\vfill

\end{document}